\documentclass[letterpaper,10pt,journal,twoside]{class/ieeeconf}

\IEEEoverridecommandlockouts
\usepackage{amsmath}
\usepackage{amssymb}
\usepackage{booktabs}
\usepackage{bm}
\usepackage{cite}
\usepackage{graphicx}
\usepackage{multirow}
\usepackage{url}
\usepackage{soul}
\usepackage[dvipsnames]{xcolor}
\usepackage[capitalize,noabbrev]{cleveref}
\usepackage{pgfplots}
\usepackage{tikz}
\usepackage{array}
\usepackage[table]{xcolor}
\usepackage[utf8]{inputenc}
\usepackage{amssymb}
\usepackage{amsfonts}
\usepackage{comment}
\usepgfplotslibrary{groupplots}
\usetikzlibrary{plotmarks}
\pgfplotsset{compat=1.16}
\usepackage{colortbl} 
\usetikzlibrary{shapes.geometric, positioning, calc, arrows.meta}
\usepackage{pifont}
\newcommand{\xmark}{\ding{55}}%

\graphicspath{{./}{../}{media/}{../media/}}

\newcommand{\ours}{\texttt{GloVLA}}

\newif\ifanon
\anonfalse          

\title{\LARGE \bf
\ours{}: Let Geometry Move and Local VLA Interact for Robust
Object-Centric Manipulation in Unstructured Environments}

\ifanon
  \author{Anonymous Authors}
\else
  \author{Truong Thanh Nguyen$^{1,\dagger}$, Huy Hoang Nguyen$^{2,\dagger,*}$,
  Ha Anh Nguyen$^{3}$, Binh Khanh Dinh$^{1}$, Ngo Anh Vien$^{1,4}$,
  Duy Nguyen Ho Minh$^{5,6,7}$, Minh Nhat Vu$^{1,4}$, and Ngan Le$^{8}$%
  \thanks{$^{\dagger}$Equal contribution.
  $^{*}$Corresponding author: {\tt\small Huy-hoang.nguyen@ait.ac.at}.
  $^{1}$VinRobotics, Vietnam;
  $^{2}$Austrian Institute of Technology, Vienna, Austria;
  $^{3}$Hanoi University of Science and Technology, Vietnam;
  $^{4}$Center for AI Research, VinUniversity, Vietnam;
  $^{5}$German Research Center for Artificial Intelligence (DFKI), Germany;
  $^{6}$University of Stuttgart, Germany;
  $^{7}$International Max Planck Research School for Intelligent Systems
  (IMPRS-IS), Germany;
  $^{8}$University of Arkansas, Fayetteville, AR, USA.
  This work has been submitted to the IEEE for possible publication.
  Copyright may be transferred without notice, after which this version may
  no longer be accessible.}%
  }
\fi
\begin{document}
\maketitle

\begin{abstract}
Vision-language-action (VLA) models have shown promising generalization for language-conditioned robot manipulation, but deploying them in unstructured environments remains challenging. 
A single end-to-end VLA policy must simultaneously solve long-range transport of the end effector to task-relevant regions and short-horizon, contact-rich interaction upon arrival.
This formulation is inefficient and brittle: small visual shifts, distractors, clutter, occlusions, or unfavorable initial gripper poses can push the policy outside the local state distribution in which it was trained, leading to task failure.
We introduce \textbf{\ours{}}, a hybrid framework that explicitly separates object-centric manipulation into two complementary regimes: a geometric transport controller moves the end-effector into interaction-centric handoff regions, and local VLA policies handle only the short-horizon interaction phases.
\ours{} is model-agnostic and can be integrated with different VLA backbones with no additional demonstrations and no changes to the action space or success predicate.
Experiments on standard LIBERO and LIBERO-Plus Object tasks together with a newly introduced \textit{LIBERO-Challenge} benchmark and real-world unstructured settings with clutter, distractors, illumination changes, visual shifts, and obstruction show that \ours{} improves task success and substantially lowers VLA inference cost compared with full end-to-end VLA execution.
On LIBERO-Challenge, full-trajectory GR00T~N1.6 execution degrades to $20.9\%$ average success while \ours{} retains $88.5\%$; on a physical UR10e, overall success improves from $35.6\%$ to $90.0\%$ while mean inference time is more than halved. Videos and additional results are available at \url{https://glovla-project.github.io/}.

\end{abstract}


\section{Introduction}

Vision-language-action (VLA) models have emerged as a promising paradigm for general-purpose robot manipulation. By mapping visual observations and language instructions directly to robot actions, models such as RT-1, RT-2, PaLM-E, OpenVLA,  $\pi_0$, $\pi_{0.5}$, $\pi_0$-FAST, $\pi_{0.6}$, GR00T can transfer semantic and visuomotor knowledge across objects, tasks, and embodiments~\cite{brohan2022rt,zitkovich2023rt2,driess2023palme,kim2024openvla,octo2024octo,o2024open,black2024pi0,black2025pi,pertsch2025fast,bjorck2025gr00tn1,intelligence2025pi}. 
%
However, real-world manipulation in unstructured environments requires more than semantic object recognition and action prediction. It must reach the workspace safely, avoid clutter and obstacles, maintain feasible kinematics, and perform the final contact-rich interaction. This requires two different regimes: \textit{global geometric approach}, where collision-free motion and reachability dominate, and \textit{local semantic manipulation}, where language grounding, object affordance, and fine-grained visuomotor control dominate. Most existing VLAs learn both regimes using a single policy.
While this design is elegant, it forces the learned policy to solve long-horizon geometric transport and local contact-rich manipulation using the same action distribution. As a result, the VLA may spend many steps on free-space motion before reaching the object, increasing computational cost and data requirements and accumulating errors over long rollouts. More importantly, unfavorable initial gripper poses, clutter, distractors, occlusions, illumination changes, and visual shifts can push the robot into states far from the demonstrations (demos) used to train the policy. Recent research shows that VLA policies remain brittle under such out-of-distribution visual and semantic variation, especially during long-horizon execution~\cite{din2025vision,zhao2025unveiling,neary2025improving}.

\begin{figure}[!t]
  \centering
  \includegraphics[width=\linewidth]{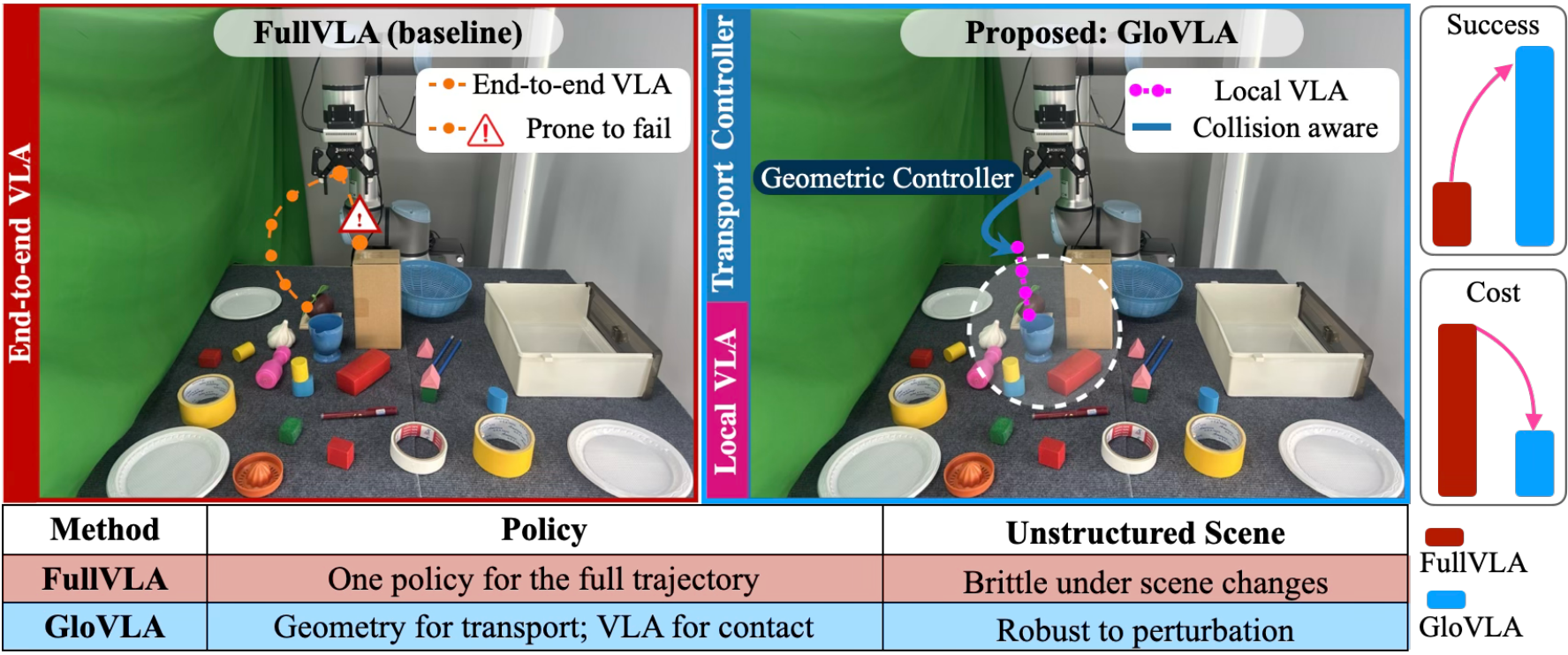}
  \caption{Comparison between existing VLA~\cite{brohan2022rt,zitkovich2023rt2,driess2023palme,kim2024openvla,octo2024octo,o2024open,black2024pi0,black2025pi,pertsch2025fast,bjorck2025gr00tn1} (left) and \ours{} (right). Existing VLAs typically rely on a single policy to handle both long-range geometric transport and local contact-rich manipulation, making them brittle to unstructured environments, data inefficiency, and high inference cost. In contrast, \ours{} factorizes manipulation into two complementary phases: a geometric controller handles transport, while the VLA focuses solely on local, semantically contact-rich manipulation. This decomposition improves robustness, data efficiency, and inference efficiency while keeping the advantages of VLA.}
  \vspace{-0.2in}
  \label{fig:teaser}
\end{figure}


The dominant response keeps the single-policy interface intact and tries to make it robust from the outside: scaling data and model capacity, adding test-time search or verification on top of the same monolithic action distribution, or post-training the policy interactively~\cite{neary2025improving,tan2025interactive,salamatian2026value}. This paper takes a complementary view: \textbf{a VLA should not be responsible for every part of a manipulation trajectory}. Classical motion planners are reliable and efficient for collision-free motion in free space when target poses and obstacles are available. At the same time, VLAs are powerful for language-conditioned local manipulation, where perception, semantics, and contact matter. 
The question is therefore not whether to use geometric control or a VLA, but how to factorize a manipulation trajectory between them. Our key insight is
that a VLA need not control the entire trajectory to retain its semantic and visuomotor advantages: repeatable free-space transport can be delegated to a simple geometric controller, and, as our experiments show, even a fixed deterministic handoff boundary suffices to recover most of the robustness lost by end-to-end VLA execution.

Based on this insight, we introduce \textbf{\ours{}} (\textbf{G}eometry +
\textbf{Lo}cal VLA), a hybrid manipulation framework for unstructured environments: assign repeatable transport to a geometric controller and reserve learned VLA control for local, contact-sensitive interactions. Given a language instruction, current observation, and robot state, \ours{} identifies the target object referenced by the instruction and localizes it (from simulator state in simulation and via open-vocabulary segmentation on the real robot). A transport controller then drives the end effector to a handoff position defined by a fixed object-centric offset, after which the VLA takes over in closed loop to complete the interaction, as shown in Figure~\ref{fig:teaser}. This design has three advantages: (i) it improves robustness by ensuring that the VLA starts from a local state where it is more likely to behave reliably; (ii) it reduces computation by shortening the VLA-controlled horizon and avoiding unnecessary VLA inference during free-space transport; (iii) it improves modularity: \ours{} can be wrapped around existing VLA backbones without architectural modification.

The standard LIBERO benchmark~\cite{liu2023libero} provides diverse object-centric manipulation tasks but does not systematically test the visual and geometric shifts that often destabilize VLAs. LIBERO-Plus
~\cite{fei25libero-plus} addresses several isolated factors, including camera, lighting, background, and object layout, but does not explicitly study approach-corridor obstruction or graded compositions of multiple perturbations. We therefore introduce \textbf{LIBERO-Challenge}, a controlled extension of LIBERO Object with three difficulty levels: \textit{easy} uses one perturbation, \textit{medium} combines two to
three, and \textit{hard} combines four to five
(Fig.~\ref{fig:challenge_scenes}). All scenes preserve the official initial-state distribution and success predicate, enabling controlled analysis of semantic, geometric, visual, and compositional failures.

Experiments with $\pi_0$, $\pi_{0.5}$, GR00T N1.6, and GR00T N1.7 show that \ours{} consistently improves or preserves performance on standard LIBERO Object and LIBERO-Plus. The advantage becomes substantially larger under distribution shift on LIBERO-Challenge and under matched source-demo budgets. On a physical UR10e robot, success improves further while inference time is reduced by more than half.

Our main contributions are:
\begin{itemize}
    \item We introduce \textbf{\ours{}}, a simple, effective, and efficient model-agnostic framework that combines geometric transport control with closed-loop VLA control for object-centric pick-and-place without modifying the VLA architecture.

    \item We construct \textbf{LIBERO-Challenge}. This compositional robustness benchmark complements LIBERO and LIBERO-Plus by evaluating manipulation under clutter, distractors, obstructions, illumination variations, visual shifts, and their graded combinations.

    \item We show that \ours{} improves success rate, robustness, and demo efficiency while reducing VLA inference cost compared with full end-to-end VLA execution, in both simulation and physical robot.
\end{itemize}

\section{Related Work}

\noindent\textbf{Vision-language-action policies.}
Large-scale robot policies have progressed from demo-scale imitation learning toward generalist models that condition on language and visual observations \cite{brohan2022rt,zitkovich2023rt2,driess2023palme,kim2024openvla,octo2024octo,o2024open}.
In addition, $\pi_0$ and $\pi_{0.5}$ use flow-based action generation and broad co-training to improve real-world
generalization~\cite{black2024pi0,black2025pi}; FAST improves action tokenization for high-frequency VLA control~\cite{pertsch2025fast}; and GR00T N1 combines vision-language reasoning with diffusion-transformer action generation~\cite{bjorck2025gr00tn1}.
These systems demonstrate impressive flexibility, but the standard fine-tuning recipe still asks a single learned controller to model both long-range approach and local contact behavior, regardless of the backbone architecture or training scale.
\ours{} is complementary: it can use any of these VLA backbones as local skill policies while removing predictable global transport from the learned action distribution.

\noindent\textbf{VLA robustness and failure analysis.}
Existing studies report persistent sensitivity to visual and semantic distribution shift in VLA execution \cite{din2025vision,zhao2025unveiling}.
Recent methods respond by adding model-based search or verification on top of the same action distribution at test time~\cite{neary2025improving}, by interactively post-training the policy on its own failure cases~\cite{tan2025interactive}, or by learning value functions to plan and search over candidate VLA rollouts~\cite{salamatian2026value}.
These approaches accept the monolithic controller as fixed and add machinery around it.
\ours{} instead removes the source of fragility for the sub-problem where it is avoidable: rather than making full-trajectory transport more robust, it eliminates learned transport from the object-centric portion of the task altogether, leaving the VLA to specialize in the interaction phase where its visual and contact reasoning are genuinely necessary.

\noindent\textbf{Hybrid planning and learned control.}
Classical motion planners and trajectory optimizers provide strong geometric reasoning, while learned policies provide visual feedback and contact-rich behavior; combining the two has long been studied in task and motion planning~\cite{garrett2021integrated}. Sampling- and optimization-based planners such as CHOMP and the systems built around MoveIt and cuRobo support collision-aware motion
generation~\cite{zucker2013chomp,coleman2014moveit,sundaralingam2023curobo}; any of these systems could be adopted as the transport controller in \ours{} when collision-aware transport is required. A separate family of systems uses large language models to decompose
long-horizon tasks into subgoals, skills, or spatial constraints, typically
grounding each subgoal in learned affordances or
keypoints~\cite{ichter2023saycan,liang2023code,huang2023voxposer,shah2023lm,huang2022inner,fang2024moka,huang2024rekep}.
Our focus is narrower, lower-level, and deliberately practical. Rather than an LLM sequencing semantic subgoals, we use a single deterministic geometric handoff split for an object-centric task, allowing the transport controller to solve the transport subproblem directly. At the same time, the VLA is invoked only where visual and contact reasoning are needed.

\section{Methodology}
\label{sec:method}

\begin{figure}[!t]
    \centering
    \resizebox{\linewidth}{!}{\input{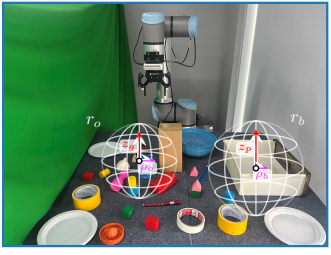}}
    \caption{\ours{} spatial parameterization diagram and corresponding experimental parameter table.}
    \label{fig:method}
\end{figure}

\ours{} follows four design principles: (i) it uses only the original full-task demos, avoiding additional human supervision; (ii) the local VLA policies retain the original observation--action interface and success predicate, while the transport controller uses object-centric state estimates for transport; (iii) the controller--policy boundary is fixed and deterministic, introducing no learned component beyond the phase policies themselves; and (iv) the factorization is backbone-agnostic and applies to any VLA with the same control interface.

\vspace{-1em}
\subsection{Problem Formulation}
\label{sec:problem}

We consider language-conditioned manipulation as a finite-horizon, partially observed sequential decision problem: the policy acts from RGB observations and proprioception, which do not constitute a Markov state. At time $t$, the robot receives 
    $o_t = (I_t^{\mathrm{ext}}, I_t^{\mathrm{wrist}}, s_t, \ell)$
: the external and wrist RGB images, the proprioceptive state, and the language instruction.
Following the LIBERO action format, each action is a 7D OSC-pose command,
\begin{equation}
    \mathbf{a}_t = (\Delta x,\Delta y,\Delta z,\Delta \phi,\Delta \theta,\Delta \psi,g)
    \in \mathbb{R}^7,
\label{eq:action-space}
\end{equation} 
where the first six dimensions are end-effector pose deltas and
$g \in [0,1]$ is the gripper command ($1$ commanding open). All policies are chunked: queried at time $t$, a policy outputs
$\mathbf{A}_t=(\mathbf{a}_t,\dots,\mathbf{a}_{t+H-1})$, of which the first $n_{\mathrm{act}} \le H$ actions are executed before re-querying ($H$, $n_{\mathrm{act}}$ in Sec.~\ref{sec:setup}).

To factorize execution, let $\mathbf{c}_o, \mathbf{c}_b \in \mathbb{R}^3$ denote the true target-object and basket centers, and let $\hat{\mathbf{c}}_o, \hat{\mathbf{c}}_b$ be the center estimates available to the controller ($\hat{\mathbf{c}}_j = \mathbf{c}_j$ in simulation; obtained from open-vocabulary segmentation on the real robot). On the real robot, $\hat{\mathbf{c}}_j$ is used in place of $\mathbf{c}_j$ in Eq.~\eqref{eq:extraction-regions}. We define two \emph{extraction regions},
\begin{equation}
    \mathcal{R}^{\mathrm{clip}}_{j} = \{\mathbf{p} \in \mathbb{R}^3 : \|\mathbf{p}-\mathbf{c}_j\|_2 \leq r_j\},\qquad j\in\{o,b\},
\label{eq:extraction-regions}
\end{equation}
where $r_o$ and $r_b$ are the grasp- and place-region radii. The extraction regions carve grasp and place clips out of the source
demos (Sec.~\ref{sec:extraction}) and thereby delimit the positional support on which the local policies are trained. They are deliberately distinct from the execution-time handoff sets $\mathcal{H}_j$ of Sec.~\ref{sec:controller}, which are centered at the controller targets rather than at the object centers; Sec.~\ref{sec:switching} gives a condition under which every handoff lands inside the corresponding extraction region.

The learned components are two independently trained local policies i.e., $\pi_g(\mathbf{A}_t \mid o_t, \ell_g)$, $\pi_p(\mathbf{A}_t \mid o_t, \ell_p),$ fine-tuned separately on the grasp and place clips $\mathcal{D}_g$ and $\mathcal{D}_p$ (Sec.~\ref{sec:extraction}), where $\ell_g$ (``pick up the \emph{target object}'') and $\ell_p$ (``place it in the basket'') are fixed phase instructions derived from $\ell$. The two policies share the backbone architecture and the observation--action interface, but not parameters. The transport controller moves the robot into the handoff sets, as described in
Sec.~\ref{sec:controller}.

\vspace{-1em}
\subsection{Sphere-Conditioned Demonstration Extraction}
\label{sec:extraction}
\noindent

The $i^{\text{th}}$ source demo is
$\tau_i=\{(o_t,\mathbf{a}_t,\mathbf{p}_t^{ee},\mathbf{G}_t)\}_{t=0}^{T_i-1}$,
where $T_i$ is the trajectory length, $\mathbf{p}_t^{ee}$ is the end-effector position and
$\mathbf{G}_t\in\mathbb{R}^2$ the observed finger-joint positions
(orientation is unused during extraction). Extraction uses three scalars:
the distances $d_j(t)=\|\mathbf{p}_t^{ee}-\mathbf{c}_j\|_2$, $j\in\{o,b\}$;
the commanded gripper coordinate $g_t\in[0,1]$ of $\mathbf{a}_t$
(Eq.~\eqref{eq:action-space}); and the aperture proxy
$\bar{G}_t=\max_k|[\mathbf{G}_t]_k|$ over the two finger joints $k$.
The gripper is \emph{commanded open} if
$O^{\mathrm{cmd}}_t\triangleq\mathbf{1}[g_t>0.5]$ and \emph{observed open}
if $O^{\mathrm{obs}}_t\triangleq\mathbf{1}[\bar{G}_t\ge 0.02\,\mathrm{m}]$;
closed states correspond to the indicator taking value $0$. Clips are
half-open index ranges, $\tau_i[t_0{:}t_1)=\{(\cdot)_t\}_{t=t_0}^{t_1-1}$;
we adopt $\min\emptyset=+\infty$ and clamp all clip endpoints to $T_i$.

The grasp clip begins at the first entry into the object region with the
gripper open by both signals, ensuring an unambiguous pre-grasp state,
\begin{equation}
    t_g^{0}=\min\{t : d_o(t)\le r_o
    \wedge O^{\mathrm{cmd}}_t=1
    \wedge O^{\mathrm{obs}}_t=1\}.
    \label{eq:grasp-start}
\end{equation}
The first subsequent close command is
\begin{equation}
    t_g^{c}=\min\{t\ge t_g^{0}: O^{\mathrm{cmd}}_t=0\},
    \label{eq:grasp-close}
\end{equation}
and the clip ends at the first exit from the object region,
\begin{equation}
    t_g^{1}=\min\{t\ge t_g^{c}: d_o(t)>r_o\}+1,
    \label{eq:grasp-end}
\end{equation}
so that $\tau_i[t_g^0{:}t_g^1)$ includes the exit step, and covers the
remainder of the trajectory if the end effector never exits the region.

The place clip starts after the grasp clip, when the end effector enters
the basket region while holding the object,
\begin{equation}
    t_p^{0}=\min\{t\ge t_g^{1}: d_b(t)\le r_b
    \wedge (O^{\mathrm{cmd}}_t=0 \vee O^{\mathrm{obs}}_t=0)\}.
    \label{eq:place-start}
\end{equation}
Unlike Eq.~\eqref{eq:grasp-start}, the held-object test is a disjunction:
the two signals can transiently disagree during transport (e.g., actuation
lag or partial closure around thin objects), and either suffices to confirm
the grasp. With $t_p^{r}=\min\{t\ge t_p^{0}: O^{\mathrm{cmd}}_t=1\}$ the
first release command, the clip ends a short fixed window later at
$t_p^{1}=t_p^{r}+9$, so that it contains the release step and the eight
subsequent steps. The local dataset is
$\mathcal{D}_{\mathrm{local}}=\mathcal{D}_g\cup\mathcal{D}_p$ with
$\mathcal{D}_g=\{\tau_i[t_g^0{:}t_g^1)\}$ and
$\mathcal{D}_p=\{\tau_i[t_p^0{:}t_p^1)\}$; a demo is retained only if
$t_g^{0}$, $t_g^{c}$, $t_p^{0}$, and $t_p^{r}$ are all finite.



\vspace{-1em}
\subsection{Transport Controller}
\label{sec:controller}
For global transport, \ours{} uses a lightweight closed-loop Cartesian controller implemented directly in LIBERO's native 7D OSC-pose action space. We refer to this component as the \emph{transport controller}; it fills the motion-planner slot of the architecture but performs no collision checking or trajectory optimization, and the factorization is agnostic to the substitution of a full motion planner (e.g., cuRobo~\cite{sundaralingam2023curobo}) when collision-aware transport is required.
For the object- and basket-approach phases, the controller defines
handoff positions computed from the estimated centers,
\begin{equation}
    \mathbf{p}_j^{h} = \hat{\mathbf{c}}_j + \bm{\delta}_j,
    \qquad j\in\{o,b\},
    \label{eq:handoff-targets}
\end{equation}
where $\bm{\delta}_o,\bm{\delta}_b\in\mathbb{R}^3$ are fixed offsets that position the end effector near the grasp and place policy distributions. We decompose each offset into a horizontal and a vertical component: $\bm{\delta}_o = (\bm{\rho}_o,\, z_g), \bm{\delta}_b = (\bm{\rho}_b,\, z_p),$
where $\bm{\rho}_o\in\mathbb{R}^2$ is the horizontal grasp-approach offset relative to the object center, $z_g$ is the grasp approach height, $\bm{\rho}_b\in\mathbb{R}^2$ is the horizontal placement offset relative to the basket center, and $z_p$ is the place approach height. These quantities, together with the extraction radii, are the spatial parameters studied in the ablation (Sec.~\ref{sec:ablation-study}).

At each controller step, the translational action is generated by a
clipped proportional law,
\begin{equation}
   \mathbf{a}_t^{xyz}
    =
    \operatorname{clip}\!
    \left(
        S^{-1}K_p\,(\mathbf{p}_j^{h}-\mathbf{p}_t^{\mathrm{ee}}),
        \,-1,\,1
    \right),
    \label{eq:controller-law}
\end{equation}
where $\mathbf{p}_t^{ee}$ is the current end-effector position, $K_p$ is a
dimensionless proportional gain, and $S$ (in meters) converts the metric position error into LIBERO's normalized OSC command range; both are scalar (isotropic) in our implementation, and the $\operatorname{clip}$ operator bounds each translational component
independently to $[-1,1]$. The rotational action dimensions are set to zero, so the controller regulates position only and the end-effector orientation is held at its value on mode entry; accordingly, $\mathbf{p}^h_j$ is a handoff \emph{position}, not a full pose. The gripper remains open during object approach and closed during transport to the basket; grasping and releasing are performed exclusively by the local VLA policies.

The controller exits its mode when the end effector enters the handoff set
\begin{equation}
    \mathcal{H}_j=\{\mathbf{p}\in\mathbb{R}^3:\|\mathbf{p}-\mathbf{p}_j^{h}\|_2\le\varepsilon_j\},\qquad j\in\{o,b\},
    \label{eq:handoff-sets}
\end{equation}
or when its step budget is exhausted; in either case, control passes to the next mode of Sec.~\ref{sec:switching}. Here $\varepsilon_j$
 is the positional tolerance at which the controller declares the handoff complete. In our implementation $\varepsilon_o = \varepsilon_b$, with the value reported in the implementation details. Note that $\mathcal{H}_j$ (centered at the controller target $\mathbf{p}^h_j$) is distinct from the extraction region $R^{\mathrm{clip}}_j$ (centered at the object center $\mathbf{c}_j$); their relation is analyzed in Sec.~\ref{sec:switching}.

\vspace{-1em}
\subsection{Hybrid Switching Policy}
\label{sec:switching}
Because geometric transport and local VLA interaction alternate along a rollout, the resulting controller is a switched system with four modes, $m_t \in \{\mathrm{app}, \mathrm{grasp}, \mathrm{trans},
    \mathrm{place}\},$
corresponding to object approach, local grasping, basket transport, and local placement. The hybrid policy is
\begin{equation}
\small
    \pi_H(\cdot \mid o_t, m_t)
    =
    \begin{cases}
        \kappa_o(\mathbf{p}_t^{ee}), & m_t = \mathrm{app},\\
        \pi_g(\cdot \mid o_t, \ell_g), & m_t = \mathrm{grasp},\\
        \kappa_b(\mathbf{p}_t^{ee}), & m_t = \mathrm{trans},\\
        \pi_p(\cdot \mid o_t, \ell_p), & m_t = \mathrm{place},
    \end{cases}
    \label{eq:hybrid-policy}
\end{equation}
where $\kappa_j(\mathbf{p}_t^{\mathrm{ee}})=(a_t^{xyz},0,0,0,\bar{g}_j)\in\mathbb{R}^7$ applies \eqref{eq:controller-law} with target $\mathbf{p}^h_j$, zero rotational increments, and the phase-appropriate gripper command (open for $j=o$, closed for $j=b$), ($\bar{g}_o = 1, \bar{g}_b = 0$) interpreted as an atomic distribution over single actions. In the learned modes, the first $n_{\mathrm{act}}$ actions of each sampled chunk $\mathbf{A}_t$ are executed before re-querying; in the controller modes, $\kappa_j$ is evaluated at every timestep.

The mode transitions are deterministic guards,
\begin{equation}
    \begin{array}{c}
\mathrm{app}
\xrightarrow{\,\mathbf{p}_t^{ee}\in \mathcal{H}_o \,\vee\, B_{\mathrm{app}}\,}
\mathrm{grasp}
\xrightarrow{\,\mathrm{lift} \,\vee\, B_{\mathrm{grasp}}\,}
\mathrm{trans} \\[4pt]
\xrightarrow{\,\mathbf{p}_t^{ee}\in \mathcal{H}_b \,\vee\, B_{\mathrm{trans}}\,}
\mathrm{place}
\xrightarrow{\,\mathrm{success} \,\vee\, B_{\mathrm{place}}\,}
\mathrm{end}
    \end{array}
    \label{eq:mode-transitions}
\end{equation}
where $B_m=\mathbf{1}[t-t_m\geq N_m]$ denotes exhaustion of mode $m$'s step budget $N_m$ with $t_m$ the mode-entry time and $m \in \{\text{app}, \text{grasp}, \text{trans}, \text{place}\}$. Lift holds once the object has been raised, and success is the unchanged official LIBERO predicate. Episodes begin in app, and budget exhaustion never terminates them: control passes to the next mode, so a timed-out transport hands the local policy a suboptimal start rather than aborting.


\textbf{Coverage of the extraction region.}
The switching structure links the execution-time quantities
(localization error, controller tolerance, and offsets) to the
training-time extraction radius. Suppose the center estimate satisfies
$\|\hat{\mathbf{c}}_j - \mathbf{c}_j\|_2 \le \eta_j$ and the controller hands off with
$\mathbf{p}_t^{ee} \in \mathcal{H}_j$. Since $\mathbf{p}^h_j = \hat{\mathbf{c}}_j + \bm{\delta}_j$, the triangle inequality gives
\begin{equation}
    \|\mathbf{p}_t^{ee} - \mathbf{c}_j\|_2 \le \varepsilon_j + \eta_j + \|\bm{\delta}_j\|_2,
    \label{eq:coverage-bound}
\end{equation}
so every handoff position lies inside the extraction region
$R^{\mathrm{clip}}_j$, and hence inside the positional support of the local training clips, whenever $\varepsilon_j + \eta_j + \|\bm{\delta}_j\|_2 \le r_j.$
The condition above guarantees positional containment in $R^{\mathrm{clip}}_j$ only; it is not a sufficient in-distribution guarantee, since orientation, gripper state, and visual context are unconstrained. Nevertheless, it makes explicit why real-robot deployment requires bounding the segmentation-based 3-D localization error $\eta_j$, and it disciplines the choice of $\varepsilon_j$. This containment applies to handoffs triggered by $\mathcal{H}_j$-entry; a budget-exhausted transport phase may hand off outside $R^{\mathrm{clip}}_j$, which is one of the failure modes discussed in Sec.~\ref{sec:discussion}.


\section{Experiments}
\label{sec:experiments}


Our experiments are organized around the four design principles in Sec.~\ref{sec:method} and four questions:

\noindent
\textbf{RQ\#1}: Does \ours{} improve manipulation success over full end-to-end VLA execution (FullVLA)(Sec.~\ref{sec:clean_results})?
\noindent
\textbf{RQ\#2}: Does \ours{} improve robustness under unstructured-scene perturbations (Sec.~\ref{sec:robustness_results})?
\noindent
\textbf{RQ\#3}: Does \ours{} improve demo efficiency (Sec.~\ref{sec:label_efficiency})?
\noindent
\textbf{RQ\#4}: Does the simulation result transfer to a physical robot (Sec.~\ref{sec:real_robot_results})?

\vspace{-1em}
\subsection{Experimental Setup}
\label{sec:setup}
\noindent
\textbf{Benchmark and Baselines.} We instantiate the learned action-level phase with representative VLA backbones, including $\pi_0$, $\pi_{0.5}$, GR00T N1.6, and GR00T N1.7.
The detailed robustness study uses GR00T N1.6 because it provides the most complete set of trained checkpoints for the simulation and real-robot evaluations.


\begin{figure}[!t]
  \centering
  \includegraphics[width=\linewidth]{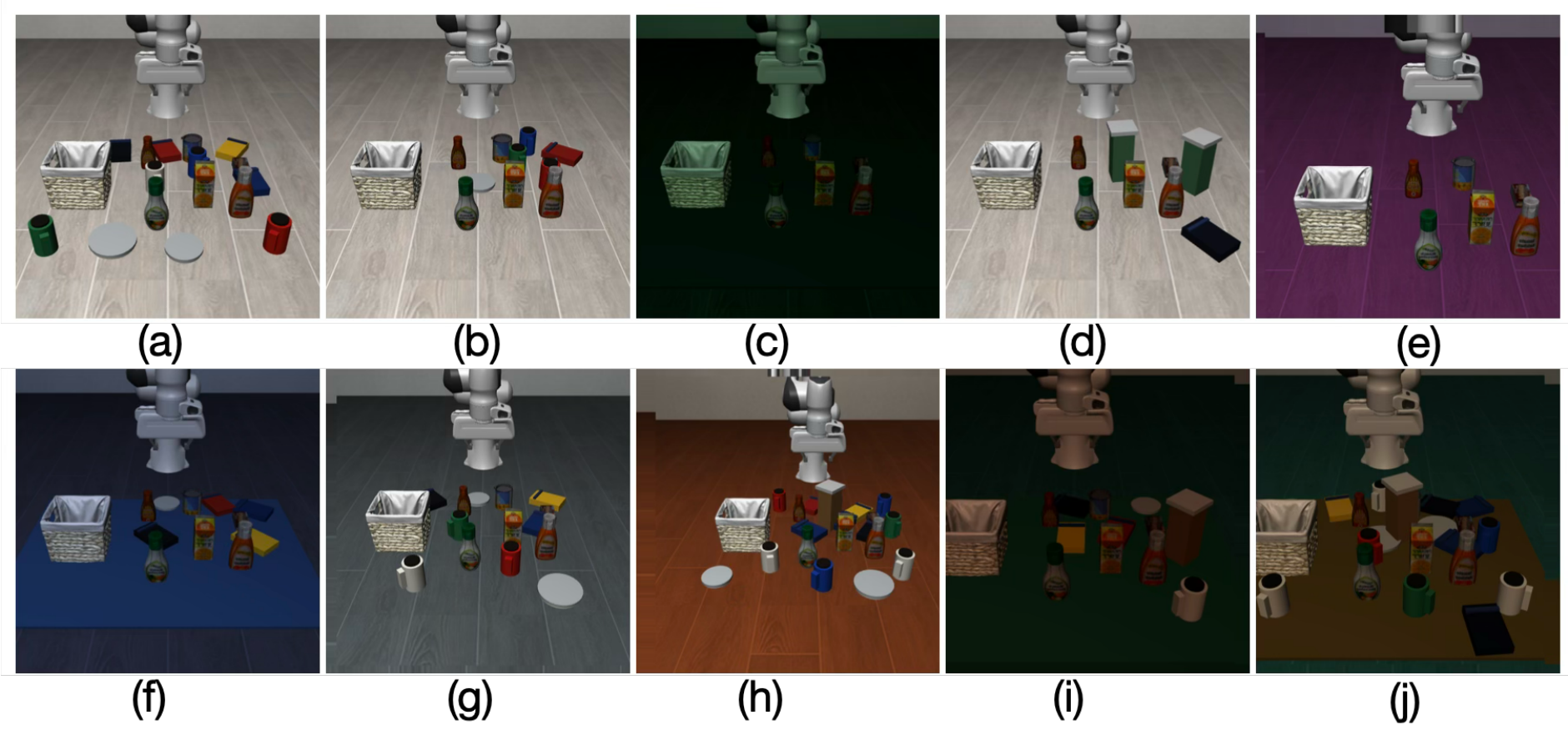}
  \vspace{-0.32in}
  \caption{Example scenes from the 50-scene \textbf{LIBERO-Challenge}. (a)--(e): \emph{easy} settings with a single perturbation (clutter, distraction, illumination, obstruction, visual shift); (f)--(g): \emph{medium} settings with 2--3 combined perturbations; and (h)--(j) \emph{hard} settings with 4--5 combined perturbations.}
\label{fig:challenge_scenes}
  \vspace{-0.1in}
\end{figure}
We evaluate on (i) standard LIBERO Object suites, (ii) standard LIBERO Plus Object suites  and (iii) our \textbf{LIBERO-Challenge} (Fig.~\ref{fig:challenge_scenes}), a stress-test benchmark derived from LIBERO Object. LIBERO-Challenge introduces five unstructured-scene perturbations: clutter, visually or semantically similar distractors, illumination shift, camera and appearance shift, and obstruction of the end-effector approach corridor. Perturbation severity is varied by changing the number of added OOD objects, lighting intensity, camera position and field of view, and scene appearance.

LIBERO-Challenge contains 50 evaluation-only scenes: 30 \textit{easy} scenes with one perturbation, 10 \textit{medium} scenes with two or three perturbations, and 10 \textit{hard} scenes with four or five. Each scene preserves the official LIBERO Object initial-state distribution and success predicate, and introduces no additional training demonstrations. Every method is evaluated on all 50 scenes using 50 episodes per scene.

\noindent
\textbf{Implementation Details.}
All simulation experiments follow the default LIBERO configuration. Policies are trained on eight NVIDIA A100 80\,GB GPUs using Adam with a learning rate of $1\times10^{-4}$. FullVLA and the local \ours{} policies use the same RGB observations, proprioceptive state, 7D OSC-pose action space, and official success predicates. \\
In simulation, object and basket centers are obtained from the environment state; on the UR10e platform, SAM~3~\cite{carion2025sam} provides their localization. The policies output chunks at each backbone's native horizon $H$ ($50$ for GR00T N1.6, $40$ for GR00T N1.7, $10$ for $\pi_0$ and $\pi_{0.5}$), of which $n_{\mathrm{act}}=8$ consecutive actions are executed per query. Evaluation is performed on an NVIDIA RTX~5090 32\, GB GPU. Parameter values are summarized in Figure~\ref{fig:method}.

\noindent
\textbf{Training and evaluation.}
FullVLA uses complete trajectories, whereas \ours{} uses paired grasp
and place clips from the same retained source demos. We report
closed-loop success under the official LIBERO predicate. 
Unless stated otherwise, each setting uses 50
episodes, organized into five independent runs of 10 episodes, reported as mean $\pm$ sample standard deviation over run-level success rates. Results with 10 episodes are reported as aggregate success only.

\vspace{-1em}
\subsection{Standard LIBERO Object Results}
\label{sec:clean_results}

\begin{table*}[t]
\centering
\setlength{\tabcolsep}{9pt}
\renewcommand{\arraystretch}{1.2}
\caption{Performance comparison between FullVLA and \ours{} across ten \textbf{LIBERO Object} tasks. \vspace{-1em}}
\label{tab:libero_object_comparison}
\resizebox{\textwidth}{!}{
\begin{tabular}{l|ll|ll|ll|ll}
\toprule
\multirow{2}{*}{\textbf{LIBERO Object Task}}
& \multicolumn{2}{c|}{\textbf{$\pi_0$}}
& \multicolumn{2}{c|}{\textbf{$\pi_{0.5}$}}
& \multicolumn{2}{c|}{\textbf{GR00T N1.6}}
& \multicolumn{2}{c}{\textbf{GR00T N1.7}} \\
\cmidrule{2-9}
& FullVLA & \textbf{\ours} & FullVLA & \textbf{\ours} & FullVLA & \textbf{\ours} & FullVLA & \textbf{\ours} \\
\midrule
Alphabet Soup
& 94.0$\pm$12.0 & \textbf{100.0$\pm$0.0}(\textcolor{Green}{\textbf{$\uparrow$ 6.0}})
& \textbf{100.0$\pm$0.0} & \textbf{100.0$\pm$0.0}
& \textbf{100.0$\pm$0.0} & \textbf{100.0$\pm$0.0}
& 98.0$\pm$4.5 & \textbf{100.0$\pm$0.0}(\textcolor{Green}{\textbf{$\uparrow$ 2.0}}) \\
BBQ Sauce
& 82.0$\pm$4.5 & \textbf{90.0$\pm$0.0} (\textcolor{Green}{\textbf{$\uparrow$ 8.0}})
& \textbf{100.0$\pm$0.0} & \textbf{100.0$\pm$0.0}
& 60.0$\pm$15.8 & \textbf{90.0$\pm$8.9} (\textcolor{Green}{\textbf{$\uparrow$ 30.0}})
& 94.0$\pm$5.5 & \textbf{98.0$\pm$4.5} (\textcolor{Green}{\textbf{$\uparrow$ 4.0}})\\
Butter
& 84.0$\pm$11.4 & \textbf{100.0$\pm$0.0} (\textcolor{Green}{\textbf{$\uparrow$ 16}})
& 96.0$\pm$5.5 & \textbf{100.0$\pm$0.0}  (\textcolor{Green}{\textbf{$\uparrow$ 4.0}})
& \textbf{100.0$\pm$0.0} & \textbf{100.0$\pm$0.0}
& 98.0$\pm$4.5 & \textbf{100.0$\pm$0.0}  (\textcolor{Green}{\textbf{$\uparrow$ 2.0}})\\
Chocolate Pudding
& 92.0$\pm$4.5 & \textbf{98.0$\pm$4.5} (\textcolor{Green}{\textbf{$\uparrow$ 6.0}})
& 90.0$\pm$8.9 & \textbf{96.0$\pm$5.5} (\textcolor{Green}{\textbf{$\uparrow$ 6.0}})
& 90.0$\pm$0.0 & \textbf{100.0$\pm$0.0} (\textcolor{Green}{\textbf{$\uparrow$ 10.0}})
& 94.0$\pm$5.5 & \textbf{98.0$\pm$4.5} (\textcolor{Green}{\textbf{$\uparrow$ 4.0}})\\
Cream Cheese
& 88.0$\pm$11.0 & \textbf{98.0$\pm$4.5} (\textcolor{Green}{\textbf{$\uparrow$ 10.0}})
& 98.0$\pm$4.5 & \textbf{100.0$\pm$0.0} (\textcolor{Green}{\textbf{$\uparrow$ 2.0}})
& 60.0$\pm$14.1 & \textbf{100.0$\pm$0.0} (\textcolor{Green}{\textbf{$\uparrow$ 40.0}})
& 98.0$\pm$4.5 & \textbf{100.0$\pm$0.0} (\textcolor{Green}{\textbf{$\uparrow$ 2.0}})\\
Ketchup
& 92.0$\pm$7.5 & \textbf{100.0$\pm$0.0} (\textcolor{Green}{\textbf{$\uparrow$ 8.0}})
& \textbf{100.0$\pm$0.0} & \textbf{100.0$\pm$0.0} 
& \textbf{100.0$\pm$0.0} & \textbf{100.0$\pm$0.0}
& \textbf{100.0$\pm$0.0} & \textbf{100.0$\pm$0.0} \\
Milk
& 76.0$\pm$23.0 & \textbf{100.0$\pm$0.0} (\textcolor{Green}{\textbf{$\uparrow$ 24.0}})
& \textbf{100.0$\pm$0.0} & \textbf{100.0$\pm$0.0} 
& 96.0$\pm$5.5 & \textbf{100.0$\pm$0.0} (\textcolor{Green}{\textbf{$\uparrow$ 4.0}})
& 94.0$\pm$8.9 & \textbf{100.0$\pm$0.0} (\textcolor{Green}{\textbf{$\uparrow$ 6.0}})\\
Orange Juice
& 74.0$\pm$15.2 & \textbf{100.0$\pm$0.0} (\textcolor{Green}{\textbf{$\uparrow$ 26.0}})
& \textbf{100.0$\pm$0.0} & \textbf{100.0$\pm$0.0} 
& 96.0$\pm$5.5 & \textbf{100.0$\pm$0.0} (\textcolor{Green}{\textbf{$\uparrow$ 4.0}})
& \textbf{100.0$\pm$0.0} & \textbf{100.0$\pm$0.0}  \\
Salad Dressing
& 98.0$\pm$4.5 & \textbf{100.0$\pm$0.0} (\textcolor{Green}{\textbf{$\uparrow$ 2.0}})
& \textbf{100.0$\pm$0.0} & \textbf{100.0$\pm$0.0}
& 94.0$\pm$12.0 & \textbf{98.0$\pm$4.5} (\textcolor{Green}{\textbf{$\uparrow$ 4.0}})
& \textbf{100.0$\pm$0.0} & \textbf{100.0$\pm$0.0} \\
Tomato Sauce
& 94.0$\pm$5.5 & \textbf{100.0$\pm$0.0} (\textcolor{Green}{\textbf{$\uparrow$ 6.0}})
& 94.0$\pm$5.5 & \textbf{100.0$\pm$0.0} (\textcolor{Green}{\textbf{$\uparrow$ 6.0}})
& 92.0$\pm$7.5 & \textbf{96.0$\pm$5.5} (\textcolor{Green}{\textbf{$\uparrow$ 4.0}})
& 88.0$\pm$13.0 & \textbf{100.0$\pm$0.0} (\textcolor{Green}{\textbf{$\uparrow$ 12.0}})\\
\midrule
\textbf{Average}
& 87.4$\pm$9.9 &  \textbf{98.6$\pm$0.9} (\textcolor{Green}{\textbf{$\uparrow$ 11.2}})
& 97.8$\pm$2.4 &  \textbf{99.6$\pm$0.6} (\textcolor{Green}{\textbf{$\uparrow$ 1.8}})
& 88.8$\pm$6.0 &  \textbf{98.4$\pm$1.9} (\textcolor{Green}{\textbf{$\uparrow$ 9.6}})
& 96.4$\pm$4.6 &  \textbf{99.6$\pm$0.9} (\textcolor{Green}{\textbf{$\uparrow$ 3.2}}) \\
\bottomrule
\end{tabular}}
\end{table*}

Table~\ref{tab:libero_object_comparison} compares FullVLA and \ours{} on ten standard LIBERO Object tasks across four VLA backbones. \ours{} consistently improves on or matches FullVLA across all tasks and backbones, showing that the controller-to-VLA handoff does not degrade performance, even when the original VLA is already strong. The gains are especially pronounced for weaker or less stable full-policy executions: $\pi_0$ improves from $87.4\%$ to $98.6\%$ average success rate, while GR00T N1.6 improves from $88.8\%$ to $98.4\%$. For stronger backbones that already approach saturation, such as $\pi_{0.5}$ and GR00T N1.7, \ours{} still provides additional gains, increasing average success from $97.8\%$ to $99.6\%$ and from $96.4\%$ to $99.6\%$, respectively. These results indicate that the benefit of \ours{} is not tied to a specific VLA architecture but stems from reducing the burden on the learned policy during the geometric approach phase. This result strongly supports \textbf{RQ\#1}.

\vspace{-1em}
\subsection{Robustness Under Unstructured Environments}
\label{sec:robustness_results}
\begin{table}[t]
\centering
\setlength{\tabcolsep}{6pt}
\renewcommand{\arraystretch}{1.2}
\caption{
Robustness on \textbf{LIBERO-Challenge}  \vspace{-1em}}
\label{tab:libero_object_robustness}
\resizebox{\columnwidth}{!}{
\begin{tabular}{llccc}
\toprule
\textbf{Difficulty}
& \textbf{Perturbation group}
& \textbf{Scenes}
& \textbf{FullVLA}
& \textbf{\ours{}}
\\
\midrule
\textbf{All}
& All challenge scenes
& 50
& 20.8 $_{\pm 10.0}$ 
& \textbf{88.5 $_{\pm 9.4}$} (\textcolor{Green}{\textbf{$\uparrow$ 67.6}})
\\
\midrule
\textbf{Easy}
& \textbf{Single perturbations}
& 30
& 31.4 $_{\pm 13.2}$
& \textbf{93.8 $_{\pm 6.0}$} (\textcolor{Green}{\textbf{$\uparrow$ 62.4}})
\\
& \quad Clutter
& 6
& 23.7 $_{\pm 14.1}$
& \textbf{94.5} $_{\pm 3.7}$ (\textcolor{Green}{\textbf{$\uparrow$ 70.8}})
\\
& \quad Distraction
& 6
& 13.5 $_{\pm 11.9}$
& \textbf{91.7} $_{\pm 7.1}$ (\textcolor{Green}{\textbf{$\uparrow$ 78.2}})
\\
& \quad Obstruction
& 6
& 43.2 $_{\pm 11.8}$
& \textbf{96.2} $_{\pm 3.4}$ (\textcolor{Green}{\textbf{$\uparrow$ 53.0}})
\\
& \quad Visual shift
& 6
& 36.7 $_{\pm 19.7}$
& \textbf{90.3} $_{\pm 11.0}$ (\textcolor{Green}{\textbf{$\uparrow$ 53.6}})
\\
& \quad Illumination
& 6
& 39.7 $_{\pm 8.4}$
& \textbf{96.2} $_{\pm 5.0}$ (\textcolor{Green}{\textbf{$\uparrow$ 56.5}})
\\
\midrule
\textbf{Medium}
& 2-3 perturbations
& 10
& 8.3 $_{\pm 7.9}$
& \textbf{86.5} $_{\pm 14.0}$ (\textcolor{Green}{\textbf{$\uparrow$ 78.2}})
\\
\midrule
\textbf{Hard}
& 4-5 perturbations
& 10
& 1.5 $_{\pm 2.6}$
& \textbf{74.9} $_{\pm 15.2}$ (\textcolor{Green}{\textbf{$\uparrow$ 73.4}})
\\
\bottomrule
\end{tabular}}
\end{table}

\begin{table*}[t]
\centering
\caption{
Robustness assessment on \textbf{LIBERO-Challenge} under various tasks and permutations. 
\vspace{-1em}}
\label{tab:libero_object_pertask}

\setlength{\tabcolsep}{4pt}
\renewcommand{\arraystretch}{1.2}

\footnotesize
\resizebox{\textwidth}{!}{
\begin{tabular}{lcccccccccc}
\toprule
\multirow{2}{*}{\textbf{Object task}}
& \multicolumn{2}{c}{\textbf{Clutter}}
& \multicolumn{2}{c}{\textbf{Distraction}}
& \multicolumn{2}{c}{\textbf{Obstruction}}
& \multicolumn{2}{c}{\textbf{Visual shift}}
& \multicolumn{2}{c}{\textbf{Illumination}} \\
\cmidrule(lr){2-3}\cmidrule(lr){4-5}\cmidrule(lr){6-7}\cmidrule(lr){8-9}\cmidrule(lr){10-11}
& FullVLA & \textbf{\ours{}} & FullVLA & \textbf{\ours{}} & FullVLA & \textbf{\ours{}} & FullVLA & \textbf{\ours{}} & FullVLA & \textbf{\ours{}} \\
\midrule
Alphabet soup & 26.3$\pm$20.9 & \textbf{99.7$\pm$0.8} (\textcolor{Green}{\textbf{$\uparrow$ 73.4}})& 12.0$\pm$14.4 & \textbf{100.0$\pm$0.0}(\textcolor{Green}{\textbf{$\uparrow$ 88.0}}) & 66.7$\pm$15.8 & \textbf{100.0$\pm$0.0}(\textcolor{Green}{\textbf{$\uparrow$ 33.3}}) & 54.3$\pm$19.6 & \textbf{100.0$\pm$0.0}(\textcolor{Green}{\textbf{$\uparrow$ 45.7}}) & 83.3$\pm$7.7 & \textbf{100.0$\pm$0.0}(\textcolor{Green}{\textbf{$\uparrow$ 16.7}}) \\

BBQ sauce & 3.3$\pm$4.8 & \textbf{94.7$\pm$3.0} (\textcolor{Green}{\textbf{$\uparrow$ 91.4}}) & 0.0$\pm$0.0 & \textbf{98.7$\pm$1.5} (\textcolor{Green}{\textbf{$\uparrow$ 98.7}}) & 10.0$\pm$5.7 & \textbf{86.7$\pm$7.0} (\textcolor{Green}{\textbf{$\uparrow$ 76.7}}) & 9.7$\pm$8.8 & \textbf{88.7$\pm$11.5} (\textcolor{Green}{\textbf{$\uparrow$ 79.0}}) & 4.3$\pm$2.7 & \textbf{90.3$\pm$9.1} (\textcolor{Green}{\textbf{$\uparrow$ 86.0}})  \\

Butter & 16.3$\pm$15.5 & \textbf{94.0$\pm$5.8} (\textcolor{Green}{\textbf{$\uparrow$ 77.7}}) & 5.0$\pm$5.6 & \textbf{57.3$\pm$22.9}  (\textcolor{Green}{\textbf{$\uparrow$ 52.3}}) & 33.3$\pm$17.0 & \textbf{91.7$\pm$12.7} (\textcolor{Green}{\textbf{$\uparrow$ 58.4}}) & 37.3$\pm$21.1 & \textbf{80.3$\pm$20.9} (\textcolor{Green}{\textbf{$\uparrow$ 43.0}}) & 36.3$\pm$8.5 & \textbf{95.3$\pm$4.5} (\textcolor{Green}{\textbf{$\uparrow$ 59.0}}) \\

Chocolate pudding & 24.0$\pm$17.8 & \textbf{98.0$\pm$1.8} (\textcolor{Green}{\textbf{$\uparrow$ 74.0}}) & 10.7$\pm$15.1 & \textbf{94.0$\pm$7.4} (\textcolor{Green}{\textbf{$\uparrow$ 83.3}}) & 35.0$\pm$14.4 & \textbf{98.3$\pm$2.7}  (\textcolor{Green}{\textbf{$\uparrow$ 63.3}}) & 37.3$\pm$22.7 & \textbf{88.0$\pm$9.9} (\textcolor{Green}{\textbf{$\uparrow$ 50.7}})  & 40.0$\pm$8.4 & \textbf{94.7$\pm$6.8} (\textcolor{Green}{\textbf{$\uparrow$ 54.7}}) \\

Cream cheese & 14.3$\pm$7.3 & \textbf{90.0$\pm$4.7} (\textcolor{Green}{\textbf{$\uparrow$ 75.7}}) & 20.0$\pm$9.4 & \textbf{96.0$\pm$3.3} (\textcolor{Green}{\textbf{$\uparrow$ 76.0}}) & 24.7$\pm$4.7 & \textbf{96.7$\pm$1.0} (\textcolor{Green}{\textbf{$\uparrow$ 72.0}})& 14.3$\pm$12.5 & \textbf{86.0$\pm$19.7} (\textcolor{Green}{\textbf{$\uparrow$ 71.7}}) & 29.0$\pm$6.3 & \textbf{96.3$\pm$1.5} (\textcolor{Green}{\textbf{$\uparrow$ 67.3}}) \\

Ketchup & 71.3$\pm$8.5 & \textbf{91.3$\pm$7.2}  (\textcolor{Green}{\textbf{$\uparrow$ 20.0}}) & 19.0$\pm$13.8 & \textbf{98.0$\pm$3.1}(\textcolor{Green}{\textbf{$\uparrow$ 79.0}})& 73.7$\pm$10.6 & \textbf{100.0$\pm$0.0}(\textcolor{Green}{\textbf{$\uparrow$ 26.3}}) & 28.7$\pm$27.9 & \textbf{81.0$\pm$30.9}(\textcolor{Green}{\textbf{$\uparrow$ 52.3}}) & 28.0$\pm$7.3 & \textbf{100.0$\pm$0.0}(\textcolor{Green}{\textbf{$\uparrow$ 72.0}}) \\

Milk & 15.3$\pm$13.7 & \textbf{100.0$\pm$0.0}  (\textcolor{Green}{\textbf{$\uparrow$ 84.7}}) & 0.3$\pm$0.8 & \textbf{100.0$\pm$0.0} (\textcolor{Green}{\textbf{$\uparrow$ 99.7}}) & 28.7$\pm$4.3 & \textbf{100.0$\pm$0.0} (\textcolor{Green}{\textbf{$\uparrow$ 71.3}}) & 43.7$\pm$18.0 & \textbf{100.0$\pm$0.0} (\textcolor{Green}{\textbf{$\uparrow$ 56.3}}) & 32.7$\pm$14.6 & \textbf{100.0$\pm$0.0} (\textcolor{Green}{\textbf{$\uparrow$ 67.3}}) \\

Orange juice & 11.7$\pm$14.2 & \textbf{100.0$\pm$0.0} (\textcolor{Green}{\textbf{$\uparrow$ 88.3}})& 5.3$\pm$9.4 & \textbf{99.3$\pm$1.6} (\textcolor{Green}{\textbf{$\uparrow$ 94.0}}) & 40.0$\pm$10.6 & \textbf{100.0$\pm$0.0} (\textcolor{Green}{\textbf{$\uparrow$ 60.0}}) & 27.0$\pm$27.8 & \textbf{99.7$\pm$0.8} (\textcolor{Green}{\textbf{$\uparrow$ 72.7}}) & 31.3$\pm$9.9 & \textbf{100.0$\pm$0.0} (\textcolor{Green}{\textbf{$\uparrow$ 68.7}}) \\

Salad dressing & 31.7$\pm$22.0 & \textbf{91.7$\pm$7.7}(\textcolor{Green}{\textbf{$\uparrow$ 60.0}}) & 29.7$\pm$25.6 & \textbf{90.3$\pm$17.7}(\textcolor{Green}{\textbf{$\uparrow$ 60.6}}) & 75.3$\pm$11.8 & \textbf{97.7$\pm$3.9}(\textcolor{Green}{\textbf{$\uparrow$ 22.4}}) & 53.3$\pm$20.2 & \textbf{97.7$\pm$5.7}(\textcolor{Green}{$\uparrow$ \textbf{44.4}}) & 48.7$\pm$10.6 & \textbf{90.7$\pm$22.9}(\textcolor{Green}{\textbf{$\uparrow$ 42.0}}) \\

Tomato sauce & 23.0$\pm$16.5 & \textbf{85.7$\pm$6.1} (\textcolor{Green}{\textbf{$\uparrow$ 62.7}}) & 32.7$\pm$24.6 & \textbf{83.3$\pm$13.8} (\textcolor{Green}{\textbf{$\uparrow$ 50.6}}) & 44.3$\pm$23.0 & \textbf{90.7$\pm$6.4} (\textcolor{Green}{\textbf{$\uparrow$ 46.4}}) & 61.7$\pm$18.4 & \textbf{81.3$\pm$10.9} (\textcolor{Green}{\textbf{$\uparrow$ 19.6}})& 63.3$\pm$7.9 & \textbf{94.3$\pm$5.4}(\textcolor{Green}{\textbf{$\uparrow$ 31.0}})  \\ \midrule
\textbf{Average} & 23.7$\pm$14.1 & \textbf{94.5$\pm$3.7} (\textcolor{Green}{\textbf{$\uparrow$ 70.8}}) & 13.5$\pm$11.9 & \textbf{91.7$\pm$7.1} (\textcolor{Green}{\textbf{$\uparrow$ 78.2}}) & 43.2$\pm$11.8 & \textbf{96.2$\pm$3.4} (\textcolor{Green}{\textbf{$\uparrow$ 53.0}}) & 36.7$\pm$19.7 & \textbf{90.3$\pm$11.0} (\textcolor{Green}{\textbf{$\uparrow$ 53.6}}) & 39.7$\pm$8.4 & \textbf{96.2$\pm$5.0} (\textcolor{Green}{\textbf{$\uparrow$ 56.5}}) \\ 
\bottomrule
\end{tabular}}
\end{table*}

\begin{table}[t]
\centering
\caption{Evaluation on the \textbf{LIBERO-Plus} object suite. Averages are weighted by the number of evaluation episodes per axis, following~\cite{fei25libero-plus}.
\vspace{-1em}
}
\label{tab:libero_plus_object}
\setlength{\tabcolsep}{4pt}
\renewcommand{\arraystretch}{1.08}
\footnotesize
\resizebox{\linewidth}{!}{
\begin{tabular}{l|c|ccccccc}
\toprule
\textbf{Method} & \textbf{Avg.} & \textbf{Cam.} & \textbf{Robot} & \textbf{Lang.} & \textbf{Light} & \textbf{BG} & \textbf{Noise} & \textbf{Layout} \\
\midrule
OpenVLA-OFT~\cite{openvla_oft}                  & 66.5 & 38.9 & 25.4 & 99.0 & 73.7 & 97.6 & 72.3 & 71.8 \\
$\pi_0$-FAST~\cite{pertsch2025fast}             & 72.7 & 72.0 & 27.6 & 71.5 & 71.0 & 95.2 & 93.1 & 84.5 \\
OpenVLA-OFT\textsubscript{m}~\cite{openvla_oft} & 77.1 & 70.2 & 18.1 & 98.5 & \textbf{100.0} & 91.9 & \textbf{94.1} & 77.4 \\ 
GR00T N1.6 & 76.6 & 56.8 & 47.2 & 90.4 & \textbf{100.0} & \textbf{99.6} & 76.3 & 81.6 \\ \midrule
\textbf{\ours{}} & \textbf{88.2} & \textbf{77.3} & \textbf{70.6} & \textbf{100.0} & \textbf{100.0} & \textbf{99.6} & 92.4 & \textbf{85.9} \\
\bottomrule
\end{tabular}}
\end{table}

Tables~\ref{tab:libero_object_robustness},~\ref{tab:libero_object_pertask}, and \ref{tab:libero_plus_object} evaluate GR00T N1.6 on LIBERO-Challenge and LIBERO-Plus. On LIBERO-Challenge, FullVLA drops to $20.9\%$, whereas \ours{} achieves $88.5\%$ ($+67.6$ points) and remains effective as perturbations compound, retaining $86.5\%$ and $74.9\%$ success on medium and hard scenes. The gains are consistent across all five perturbation types and object tasks, indicating that the factorization improves both geometric approach and local interaction robustness rather than overfitting to one condition. 
This trend generalizes to LIBERO-Plus, where \ours{} obtains the highest average success ($88.2\%$) and leads or matches on six of seven axes and is competitive in the noise setting. Together, the two benchmarks show complementary evidence: LIBERO-Challenge validates robustness under controlled multi-factor composition, while LIBERO-Plus confirms broad generalization across diverse distribution shifts.
These results indicate that a major weakness of full-trajectory VLA control lies in reliably reaching and preserving a suitable local interaction state under visual and geometric shifts, supporting \textbf{RQ\#2}.

\vspace{-1em}
\subsection{Data Efficiency}
\label{sec:label_efficiency}
\begin{table}[t]
\centering
\caption{\textbf{Data efficiency} on LIBERO-Challenge (task 1).  \vspace{-1em} 
}
\label{tab:label_efficiency}
\setlength{\tabcolsep}{4pt}
\resizebox{\linewidth}{!}{
\begin{tabular}{lllll}
\toprule
\multirow{2}{*}{\textbf{Settings}} & \multirow{2}{*}{\textbf{Method}} & \multicolumn{3}{c}{\textbf{Demos}}\\
\cmidrule(lr){3-5}
& & \textbf{10} & \textbf{30} & \textbf{50}\\
\midrule
\multirow{2}{*}{Clean Table-top} & FullVLA & 86.0$\pm$15.2 & 90$\pm$8.9 & 100.0$\pm$0.0 \\
& \textbf{\ours{}} & 100.0$\pm$0.0 (\textcolor{Green}{\textbf{$\uparrow$14.0}}) & 100.0$\pm$0.0  (\textcolor{Green}{\textbf{$\uparrow$10.0}})& 100.0$\pm$0.0\\  \midrule
\rowcolor{gray!15}\multicolumn{5}{l}{Unstructured Scene} \\
\multirow{2}{*}{Clutter} & FullVLA & 8.7$\pm$6.8 & 19.7$\pm$13.6 & 26.3$\pm$20.9 \\
& \textbf{\ours{}} & 23.0$\pm$10.9 (\textcolor{Green}{\textbf{$\uparrow$14.3}}) & 99.0$\pm$1.7 (\textcolor{Green}{\textbf{$\uparrow$79.3}}) & 99.7$\pm$0.8 (\textcolor{Green}{\textbf{$\uparrow$73.4}})\\
\multirow{2}{*}{Distraction} & FullVLA & 0.3$\pm$0.8 & 9.9$\pm$6.5 & 12.0$\pm$14.4 \\
& \textbf{\ours{}} & 27.7$\pm$25.7 (\textcolor{Green}{\textbf{$\uparrow$27.4}}) & 100.0$\pm$0.0 (\textcolor{Green}{\textbf{$\uparrow$90.1}}) & 100.0$\pm$0.0  (\textcolor{Green}{\textbf{$\uparrow$88.0}})\\
\multirow{2}{*}{Obstruction} & FullVLA & 15.0$\pm$13.5 & 33.1$\pm$12.2 & 66.7$\pm$15.8 \\
& \textbf{\ours{}} & 54.3$\pm$13.5 (\textcolor{Green}{\textbf{$\uparrow$39.3}}) & 100.0$\pm$0.0 (\textcolor{Green}{\textbf{$\uparrow$66.9}}) & 100.0$\pm$0.0 (\textcolor{Green}{\textbf{$\uparrow$33.3}}) \\
\multirow{2}{*}{Visual Shift} & FullVLA & 12.3$\pm$13.4 & 17.8$\pm$13.2 & 54.3$\pm$19.6 \\
& \textbf{\ours{}} & 64.0$\pm$5.9 (\textcolor{Green}{\textbf{$\uparrow$51.7}}) & 98.7$\pm$2.4 (\textcolor{Green}{\textbf{$\uparrow$80.9}}) & 100.0$\pm$0.0 (\textcolor{Green}{\textbf{$\uparrow$45.7}})\\
\multirow{2}{*}{Illumination} & FullVLA & 9.0$\pm$5.3 & 19.1$\pm$8.6 & 83.3$\pm$7.7 \\
& \textbf{\ours{}} & 56.3$\pm$10.5 (\textcolor{Green}{\textbf{$\uparrow$47.3}}) & 100.0$\pm$0.0 (\textcolor{Green}{\textbf{$\uparrow$80.9}}) & 100.0$\pm$0.0 (\textcolor{Green}{\textbf{$\uparrow$16.7}}) \\ \midrule
\multirow{2}{*}{\textbf{Average}} & FullVLA & 9.1$\pm$8.0 & 19.9$\pm$10.8 & 48.5$\pm$15.7 \\
 & \textbf{\ours{}} & 45.1$\pm$13.3 (\textcolor{Green}{\textbf{$\uparrow$36.0}}) & 99.5$\pm$0.8 (\textcolor{Green}{\textbf{$\uparrow$79.6}}) & 99.9$\pm$0.2 (\textcolor{Green}{\textbf{$\uparrow$51.4}}) \\
\bottomrule
\end{tabular}
    }
\vspace{-2em}
\end{table}
Table~\ref{tab:label_efficiency} shows that \ours{} is substantially more data-efficient under matched source-demonstration budgets. It reaches $100\%$ success on clean tasks with 10 demos and averages $99.9\%$ across unstructured conditions with 50 demos, compared with $48.5\%$ for FullVLA. These results suggest that factorization reduces the amount of behavior learned from demos: planning handles long-range geometric transport, allowing the VLA to concentrate its limited data capacity on local semantic and contact-rich interaction, strongly supports \textbf{RQ\#3}.


\vspace{-1em}
\subsection{Real-World Demonstrations}
\label{sec:real_robot_results}

\begin{table}[t]
\centering
\setlength{\tabcolsep}{9pt}
\caption{Evaluation on \textbf{real-robot} under various settings. \vspace{-1em}}
\label{tab:real_robot}
\resizebox{\linewidth}{!}{
\begin{tabular}{llccc}
\toprule
\multicolumn{1}{l}{\textbf{Perturbation}} & \multicolumn{1}{l}{\textbf{Setting}} & \textbf{Trials} & \textbf{FullVLA} & \textbf{\ours{}} \\
\midrule
\xmark & \multicolumn{1}{l}{Table-top} & 20 &  90\% & 95\% (\textcolor{Green}{\textbf{$\uparrow$ 5.0}})\\ 
\midrule
\multirow{5}{*}{{ \checkmark \quad Easy}} &
\multicolumn{1}{l}{Clutter}  & 10 & 30\% & 90\% (\textcolor{Green}{\textbf{$\uparrow$ 60.0}})\\ 
 & \multicolumn{1}{l}{Obstruction} & 10 & 40\% & 100\% (\textcolor{Green}{\textbf{$\uparrow$ 60.0}})\\ 
 & \multicolumn{1}{l}{Illumination}& 10 & 0\% & 80\% (\textcolor{Green}{\textbf{$\uparrow$ 80.0}})  \\ 
 & \multicolumn{1}{l}{Visual shift} & 10 & 40\% & 100\%  (\textcolor{Green}{\textbf{$\uparrow$ 60.0}}) \\ 
 & \multicolumn{1}{l}{Distraction} & 10 & 20\% & 100\% (\textcolor{Green}{\textbf{$\uparrow$ 80.0}})\\  \midrule
\multicolumn{1}{l}{\checkmark \quad Medium} & \multicolumn{1}{r}{2-3 perturbations}  & 10 & 10\% & 80\% (\textcolor{Green}{\textbf{$\uparrow$ 70.0}})\\ \midrule
\multicolumn{1}{l}{\checkmark \quad Hard} & \multicolumn{1}{r}{4-5 perturbations} & 10 & 0\% & 70\% (\textcolor{Green}{\textbf{$\uparrow$ 70.0}})\\
\midrule
 \multicolumn{2}{l}{\textbf{Overall}} & 90 & 35.6\% & 90.0\% (\textcolor{Green}{\textbf{$\uparrow$ 54.4}})\\
 \midrule
 \multicolumn{2}{l}{\textbf{Inference Time (s) $\downarrow$}} & -- & 60.1$\pm$19.6 & 28.2$\pm$2.0 (\textcolor{Green}{\textbf{$\downarrow$2.13$\times$}})\\
\bottomrule
 
\end{tabular}}
\end{table}

\begin{figure}[h]
\centering
\vspace{-1em}
\begin{tikzpicture}
    \matrix[column sep=2pt, row sep=4pt, inner sep=0pt, align=center] {
        
        \node[rotate=90]{\scriptsize{FullVLA}}; & 
        \node{\includegraphics[width=0.24\linewidth]{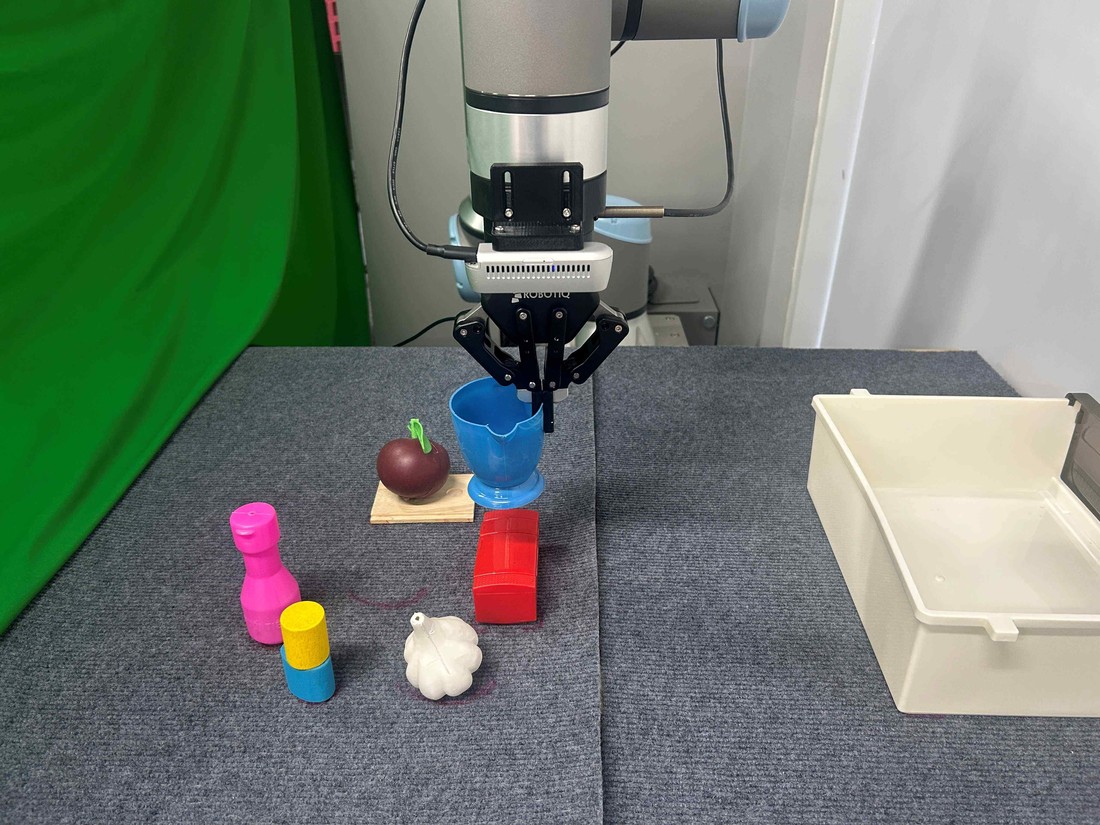}}; & 
        \node{\includegraphics[width=0.24\linewidth]{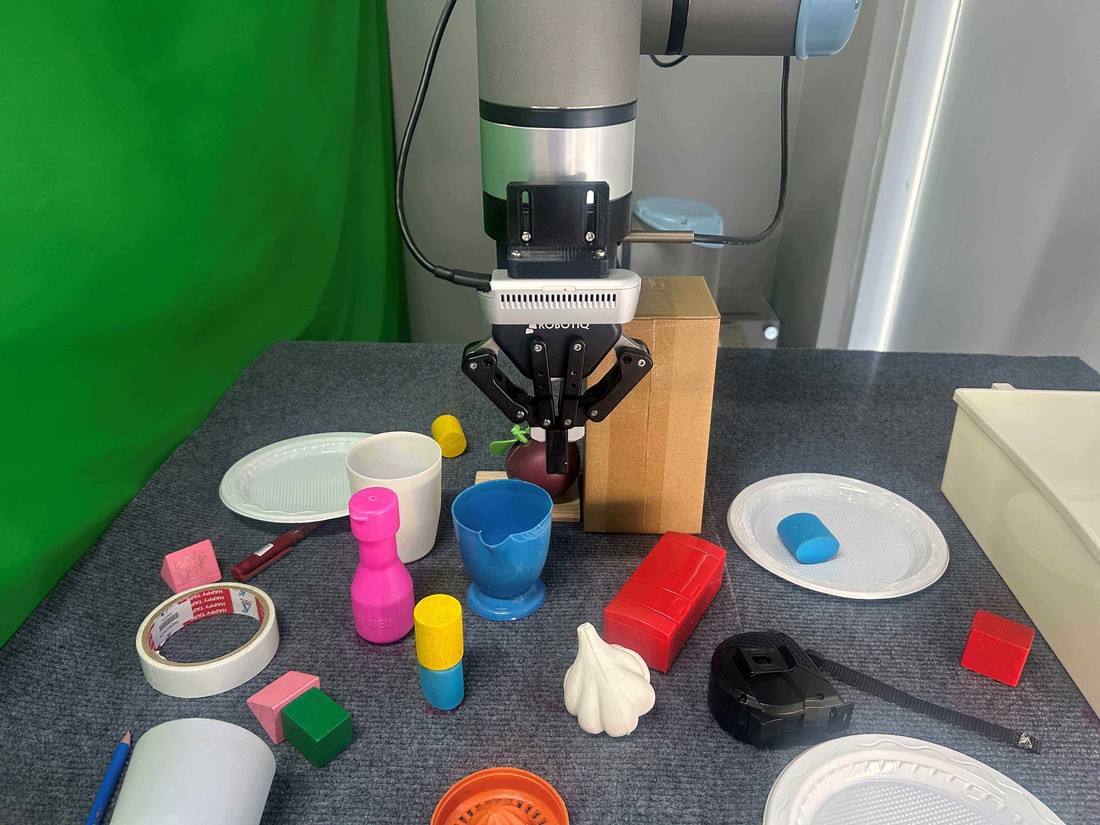}}; & 
        \node{\includegraphics[width=0.24\linewidth]{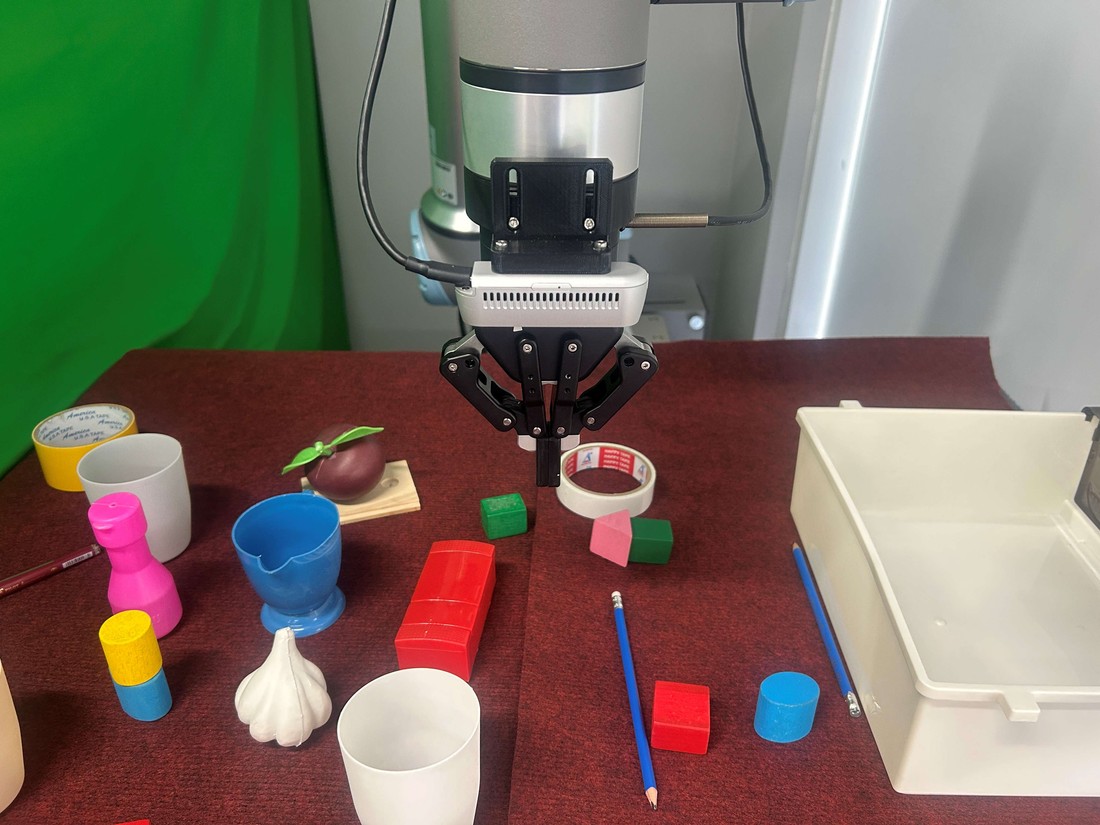}}; &
        \node{\includegraphics[width=0.24\linewidth]{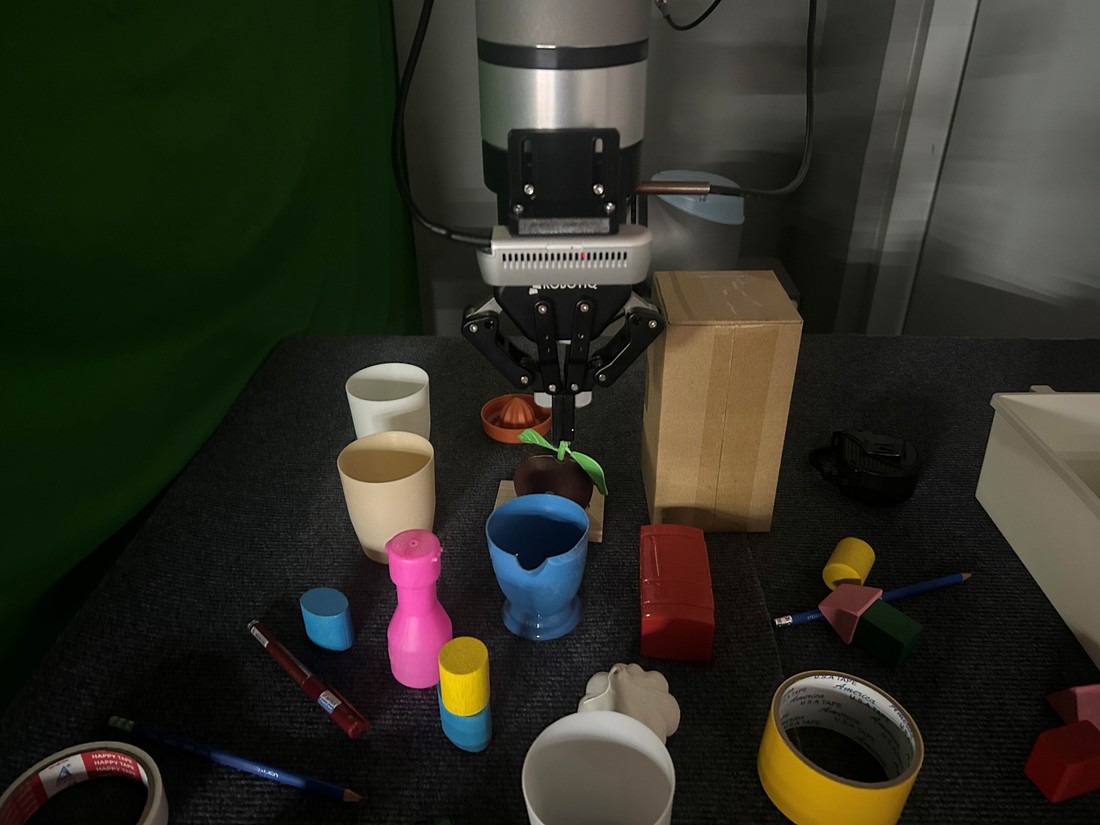}}; \\

        \node[rotate=90]{\scriptsize\textbf{\ours{}}}; & 
        \node{\includegraphics[width=0.24\linewidth]{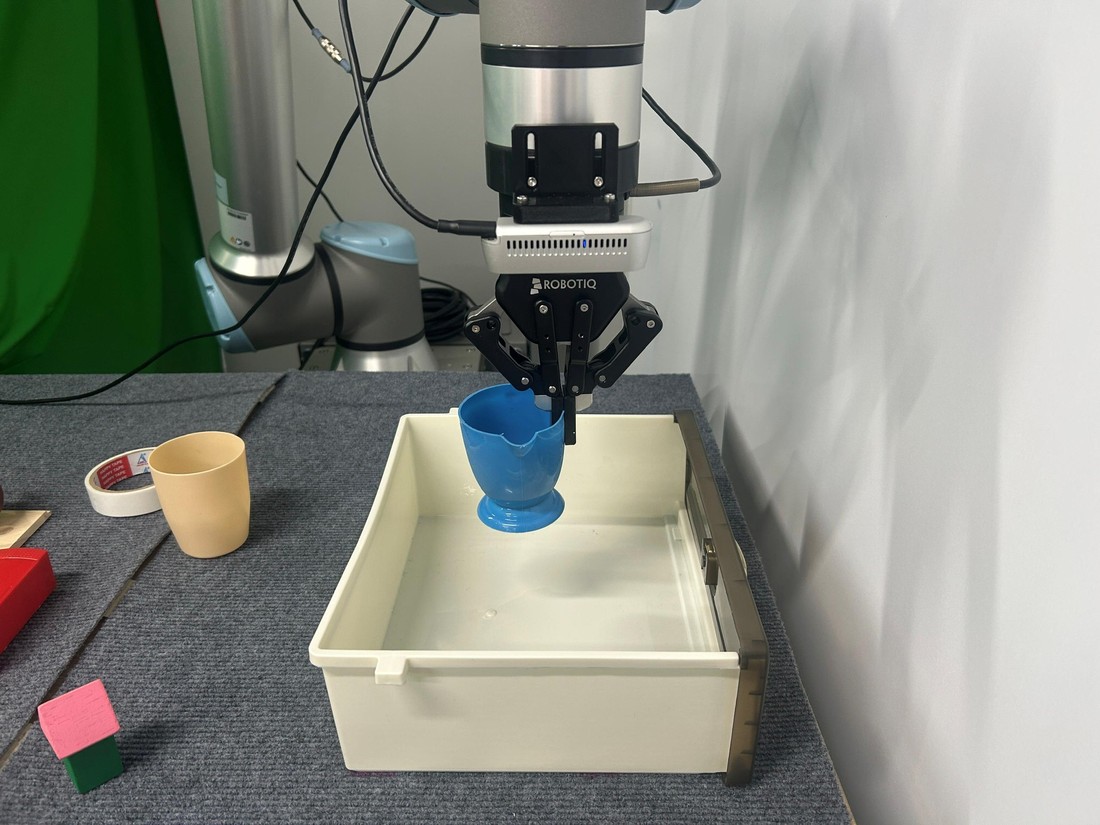}}; & 
        \node{\includegraphics[width=0.24\linewidth]{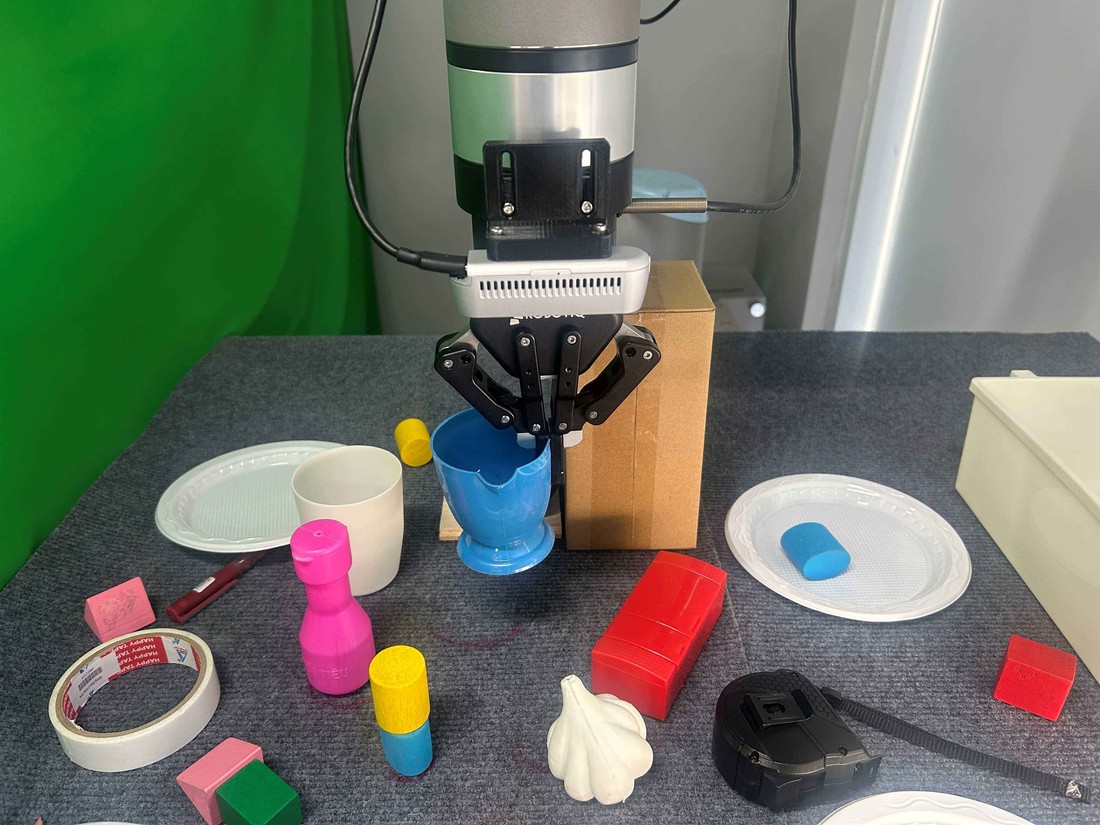}}; & 
        \node{\includegraphics[width=0.24\linewidth]{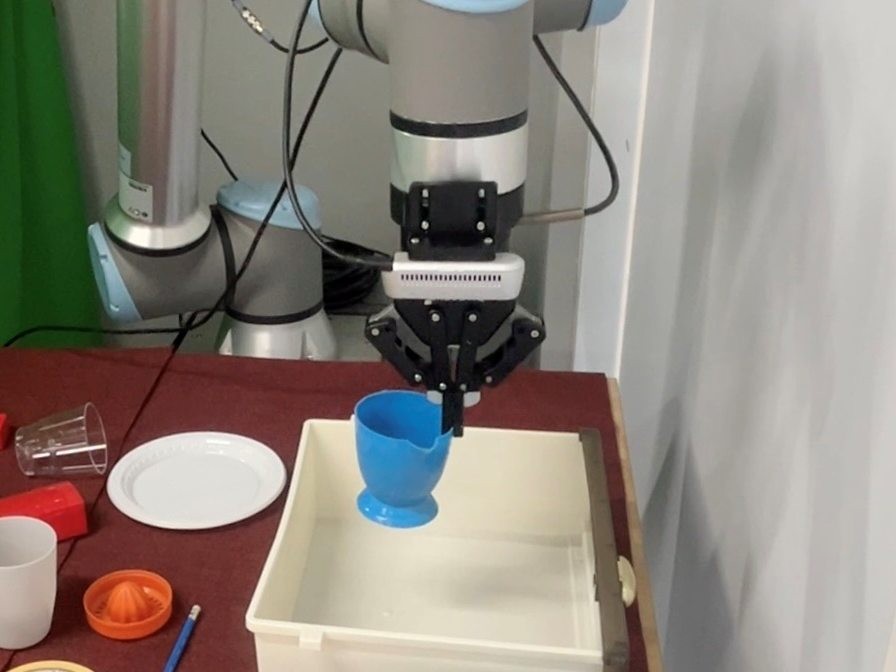}}; &
        \node{\includegraphics[width=0.24\linewidth]{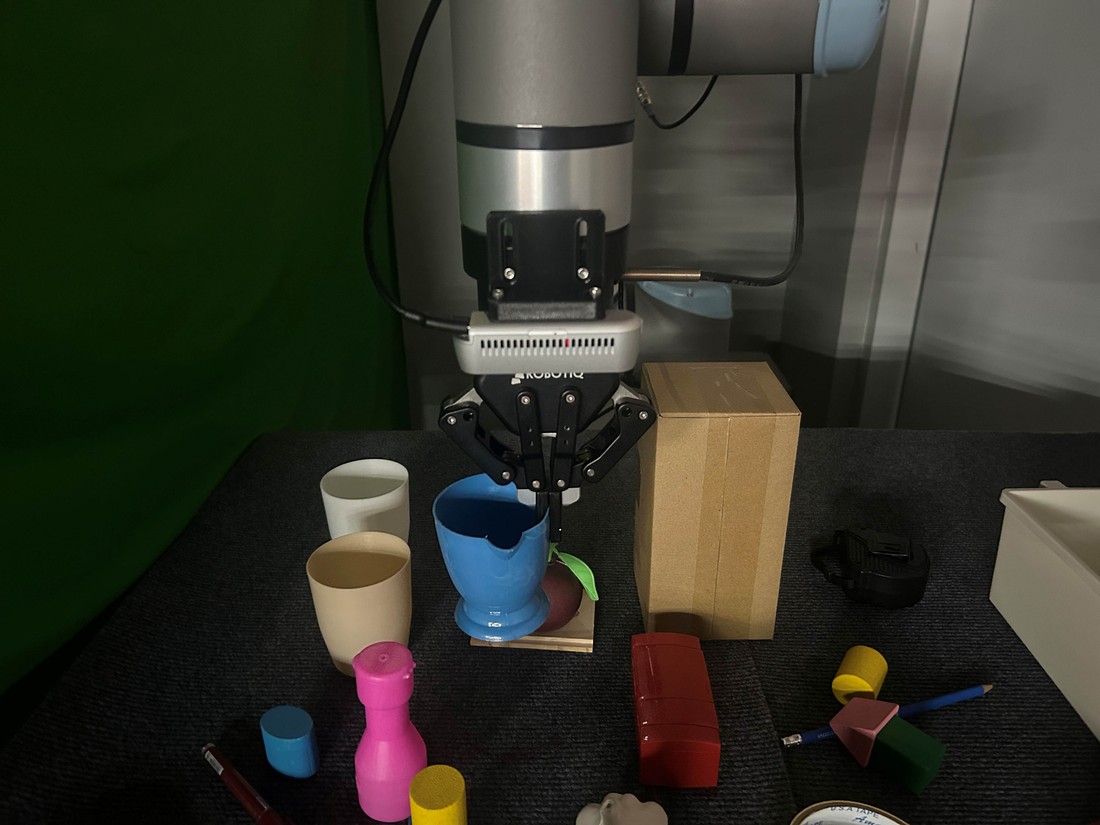}}; \\

        & 
        \node{\scriptsize{Clean Table-top}}; & 
        \node{\scriptsize{Easy}}; & 
        \node{\scriptsize{Medium}}; &
        \node{\scriptsize{Hard}}; \\
    };
\end{tikzpicture}
\caption{Qualitative comparison of success cases between FullVLA and \ours{} across varying difficulty levels in real-robot setups.}
\label{fig:qualitative}
\vspace{-1em}
\end{figure}
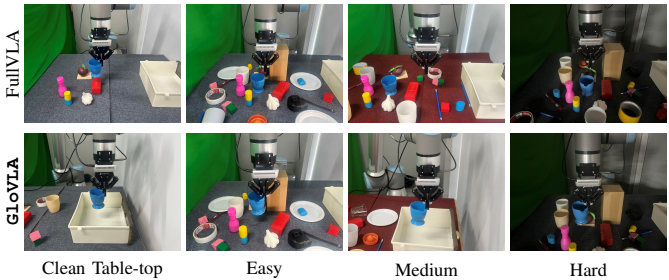

\begin{figure}[h]
\centering
\begin{tikzpicture} 
    \matrix[column sep=2pt, row sep=3pt, inner sep=0pt, align=center] {
        \node{\includegraphics[width=0.33\linewidth]{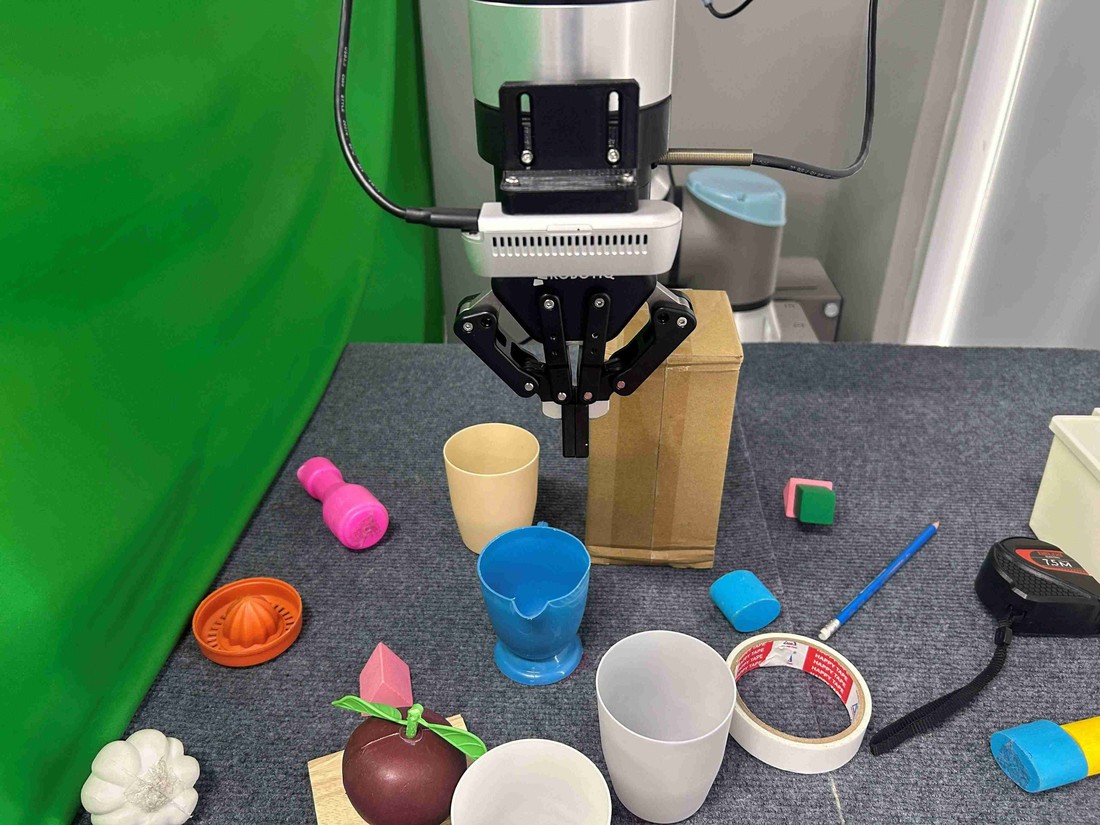}}; & 
        \node{\includegraphics[width=0.33\linewidth]{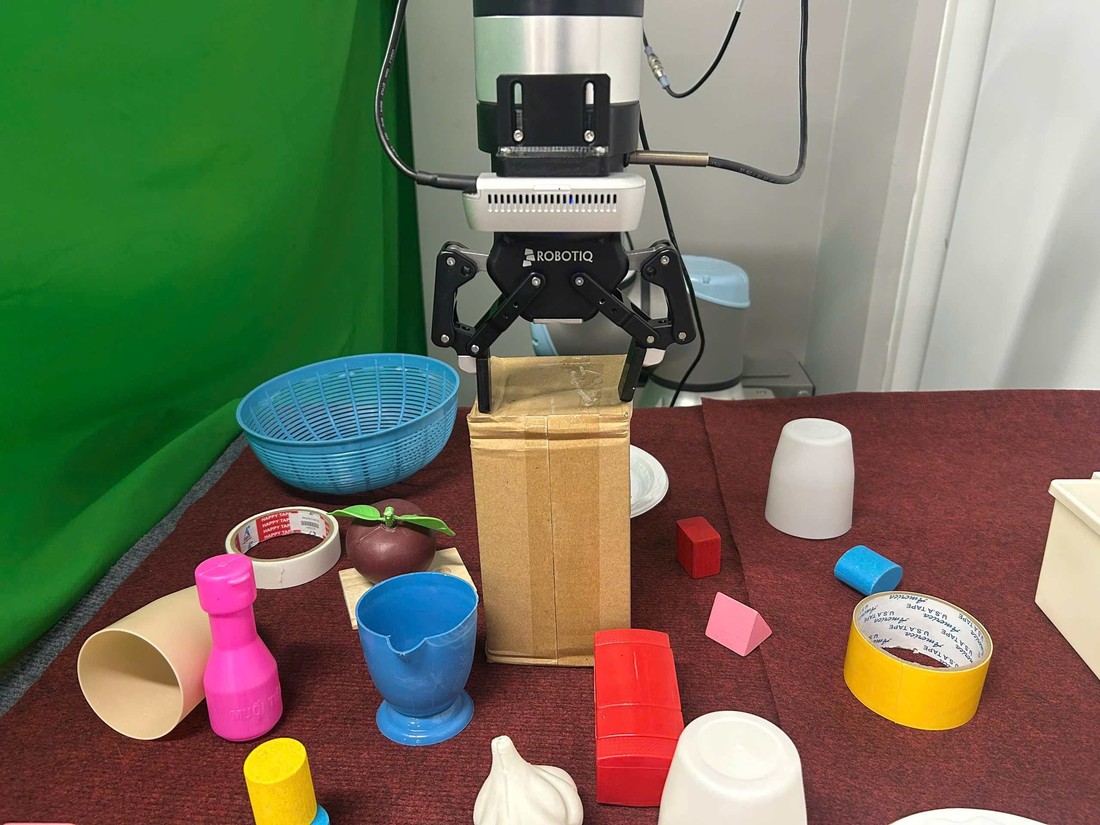}}; & 
        \node{\includegraphics[width=0.33\linewidth]{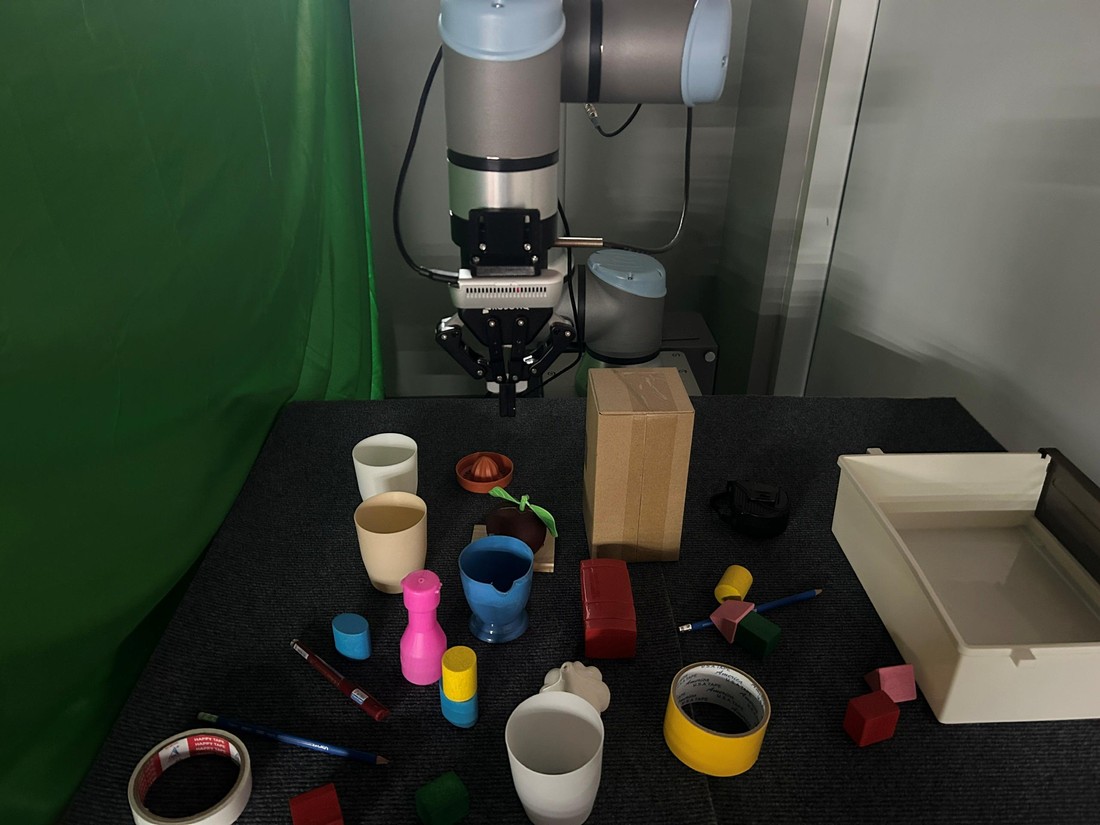}}; \\
        \node{\scriptsize{Easy}}; & 
        \node{\scriptsize{Medium}}; & 
        \node{\scriptsize{Hard}}; \\
    };
\end{tikzpicture}
\vspace{-0.3cm}
\caption{Failure cases \ours{}. Our method fails when the object center estimation is wrong, or the offset is too large.}
\label{fig:failure_cases}
\vspace{-1em}
\end{figure}
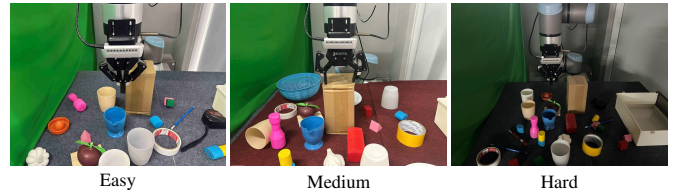


 
We further evaluate \ours{} in real-robot tabletop experiments on UR10e as in Fig.~\ref{fig:qualitative}. As shown in Table~\ref{tab:real_robot}, \ours{} improves overall success from $35.6\%$ to $90.0\%$ and reduces inference time from $60.1$\,s to $28.2$\,s. The gains are largest under illumination and distraction, while \ours{} retains $80\%$ and $70\%$ success on medium and hard compositions, respectively, compared with $10\%$ and $0\%$ for FullVLA. These results support \textbf{RQ\#4} and confirm that the robustness benefits transfer to the physical system.

\subsection{Qualitative Results}
\label{sec:qualitative}

Figure~\ref{fig:qualitative} compares FullVLA and \ours{} on four
real-robot settings of increasing difficulty: both methods succeed on
the clean table-top, but FullVLA fails under the cluttered easy scene
and continues to fail as difficulty increases (background change in
medium; combined obstruction and distraction in hard), while \ours{}
succeeds throughout, mirroring Table~\ref{tab:real_robot}.
Figure~\ref{fig:failure_cases} illustrates the three failure modes that
remain: erroneous object-center estimation places the handoff away from
the object, invoking the grasp policy outside its trained
region, and an overly large offset forces the
local policy to recover part of the transport.



\vspace{-1em}
\subsection{Ablation Study}
\label{sec:ablation-study}

\begin{figure}[t]
\centering
\resizebox{\columnwidth}{!}{
    \definecolor{cClean}{RGB}{31,119,180}
\definecolor{cClutter}{RGB}{255,127,14}
\definecolor{cDistraction}{RGB}{44,160,44}
\definecolor{cObstruction}{RGB}{214,39,40}
\definecolor{cIllumination}{RGB}{148,103,189}
\definecolor{cShift}{RGB}{140,86,75}

\begin{tikzpicture}[
    font=\sffamily,
    x=1.05cm,
    y=0.05cm 
]

\draw[->, thick]
    (0.5,75) -- (9.65,75)
    node[right] {};

\draw[->, thick]
    (0.5,75) -- (0.5,108)
    node[above] {};

\foreach \y in {80,100}
{
    \draw[dashed, gray]
        (0.5,\y) -- (9.5,\y);

    \node[anchor=east, font=\footnotesize]
        at (0.42,\y) {\y};
}

\foreach \x in {1,2,3,4,5,6,7,8}
{
    \draw (\x,75) -- (\x,73.5);
}

\node[rotate=45, anchor=north east, font=\footnotesize]
    at (1,73) {$(10,10)$};

\node[rotate=45, anchor=north east, font=\footnotesize]
    at (2,73) {$(10,15)$};

\node[rotate=45, anchor=north east, font=\footnotesize]
    at (3,73) {$(10,20)$};

\node[rotate=45, anchor=north east, font=\footnotesize]
    at (4,73) {$(10,25)$};

\node[rotate=45, anchor=north east, font=\footnotesize]
    at (5,73) {$(15,20)$};

\node[rotate=45, anchor=north east, font=\footnotesize]
    at (6,73) {$(20,20)$};

\node[rotate=45, anchor=north east, font=\footnotesize]
    at (7,73) {$(25,20)$};

\node[rotate=45, anchor=north east, font=\footnotesize]
    at (8,73) {$(30,20)$};

\node[font=\bfseries]
    at (5,48) {Configuration $(r_o,r_b)$};

\node[rotate=90,font=\bfseries]
    at (-0.25,91.5) {Success Rate (\%)};

\draw[cClean, thick]
    (1,100) -- (2,100) -- (3,100) -- (4,100) -- (5,100) -- (6,100) -- (7,100) -- (8,100);

\foreach \x in {1,2,3,4,5,6,7,8}
    \fill[cClean] (\x,100) circle (2pt);

\draw[cClutter, thick]
    (1,100) -- (2,90) -- (3,100) -- (4,100) -- (5,100) -- (6,100) -- (7,100) -- (8,100);

\foreach \x/\y in {
    1/100,2/90,3/100,4/100,
    5/100,6/100,7/100,8/100
}
{
    \filldraw[cClutter]
        (\x-0.05, \y-0.5) rectangle ++(0.1, 1.0);
}

\draw[cDistraction, thick]
    (1,90) -- (2,100) -- (3,100) -- (4,100) -- (5,100) -- (6,100) -- (7,100) -- (8,100);

\foreach \x/\y in {
    1/90,2/100,3/100,4/100,
    5/100,6/100,7/100,8/100
}
{
    \node[
        regular polygon,
        regular polygon sides=3,
        fill=cDistraction,
        inner sep=1.6pt
    ] at (\x,\y) {};
}

\draw[cObstruction, thick]
    (1,100) -- (2,100) -- (3,100) -- (4,100) -- (5,100) -- (6,100) -- (7,100) -- (8,100);

\foreach \x in {1,2,3,4,5,6,7,8}
{
    \node[
        diamond,
        fill=cObstruction,
        inner sep=1.6pt
    ] at (\x,100) {};
}

\draw[cIllumination, thick]
    (1,80) -- (2,100) -- (3,100) -- (4,100) -- (5,100) -- (6,100) -- (7,100) -- (8,100);

\foreach \x/\y in {
    1/80,2/100,3/100,4/100,
    5/100,6/100,7/100,8/100
}
{
    \draw[cIllumination, very thick]
        (\x,\y) circle (2.3pt);
}

\draw[cShift, thick]
    (1,100) -- (2,100) -- (3,100) -- (4,90) -- (5,100) -- (6,100) -- (7,100) -- (8,100);
\foreach \x/\y in {
    1/100,2/100,3/100,4/90,
    5/100,6/100,7/100,8/100
}
{
    \draw[cShift, very thick]
        (\x-0.07,\y-0.75) -- (\x+0.07,\y+0.75);

    \draw[cShift, very thick]
        (\x-0.07,\y+0.75) -- (\x+0.07,\y-0.75);
}

\begin{scope}[shift={(0.0, 112)}]

    \draw[cClean, thick] (0, 0) -- (0.3, 0);
    \fill[cClean] (0.15, 0) circle (2pt);
    \node[anchor=west, font=\scriptsize, inner sep=1pt] at (0.35, 0) {Clean};

    \draw[cClutter, thick] (1.5, 0) -- (1.8, 0);
    \filldraw[cClutter] (1.6, -0.5) rectangle ++(0.1, 1.0);
    \node[anchor=west, font=\scriptsize, inner sep=1pt] at (1.85, 0) {Clutter};

    \draw[cDistraction, thick] (3.1, 0) -- (3.4, 0);
    \node[regular polygon, regular polygon sides=3, fill=cDistraction, inner sep=1.5pt] at (3.25, 0) {};
    \node[anchor=west, font=\scriptsize, inner sep=1pt] at (3.45, 0) {Distraction};

    \draw[cObstruction, thick] (4.9, 0) -- (5.2, 0);
    \node[diamond, fill=cObstruction, inner sep=1.5pt] at (5.05, 0) {};
    \node[anchor=west, font=\scriptsize, inner sep=1pt] at (5.25, 0) {Obstruction};

    \draw[cIllumination, thick] (6.6, 0) -- (6.9, 0);
    \draw[cIllumination, very thick] (6.75, 0) circle (2.2pt);
    \node[anchor=west, font=\scriptsize, inner sep=1pt] at (6.95, 0) {Illumination};

    \draw[cShift, thick] (8.4, 0) -- (8.7, 0);
    \draw[cShift, very thick] (8.48, -0.75) -- (8.62, 0.75);
    \draw[cShift, very thick] (8.48, 0.75) -- (8.62, -0.75);
    \node[anchor=west, font=\scriptsize, inner sep=1pt] at (8.75, 0) {Visual Shift};

\end{scope}

\end{tikzpicture} 
}
\vspace{-8mm}
\caption{Ablation on the grasp and place extraction radius $r_o$, $r_b$.}
\label{fig:ablation-radius}
\vspace{-1em}
\end{figure}
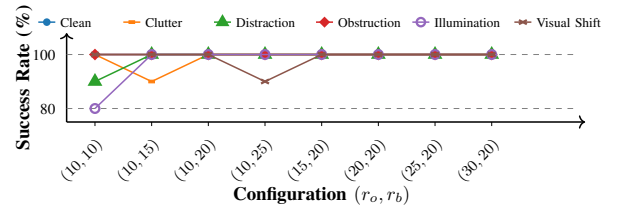

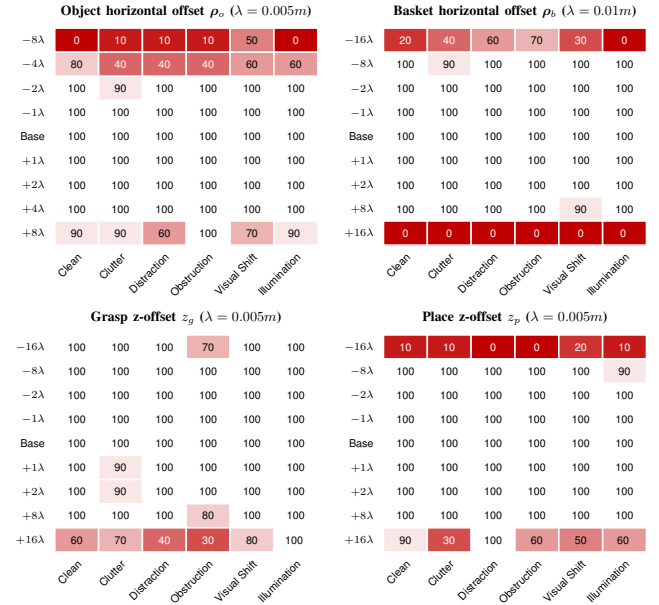
\begin{figure}[t]
\centering
\resizebox{\columnwidth}{!}{
    \begin{tikzpicture}[
    x=1.0cm, y=0.55cm,
    font=\sffamily\scriptsize
]

\definecolor{alertred}{RGB}{192, 0, 0}

\def\drawcell#1#2#3{
    \pgfmathtruncatemacro{\redval}{100 - #3}
    \fill[alertred!\redval!white] (#1-1, -#2) rectangle (#1, -#2+1);
    \draw[white, line width=1pt] (#1-1, -#2) rectangle (#1, -#2+1);
    \node at (#1-0.5, -#2+0.5) {\ifnum#3<50 \textcolor{white}{#3}\else \textcolor{black}{#3}\fi};
}

\newcommand{\drawheatmap}[3]{
    \node[font=\bfseries\small, anchor=south] at (3, 0.2) {#1};
    
    \foreach \lbl [count=\y] in {#2} {
        \node[anchor=east] at (-0.1, -\y+0.5) {\lbl};
    }
    
    \foreach \col [count=\x] in {Clean, Clutter, Distraction, Obstruction, Visual Shift, Illumination} {
        \node[rotate=45, anchor=north east, inner sep=2pt] at (\x-0.5, -9.1) {\col};
    }
    
    \foreach \vA/\vB/\vC/\vD/\vE/\vF [count=\y] in {#3} {
        \drawcell{1}{\y}{\vA}
        \drawcell{2}{\y}{\vB}
        \drawcell{3}{\y}{\vC}
        \drawcell{4}{\y}{\vD}
        \drawcell{5}{\y}{\vE}
        \drawcell{6}{\y}{\vF}
    }
}

\begin{scope}[xshift=0cm, yshift=0cm]
    \drawheatmap{Object horizontal offset $\bm{\rho}_o$ ($\lambda = 0.005m$)}
    {$-8\lambda$, $-4\lambda$, $-2\lambda$, $-1\lambda$, Base, $+1\lambda$, $+2\lambda$, $+4\lambda$, $+8\lambda$}
    {
        0/10/10/10/50/0,
        80/40/40/40/60/60,
        100/90/100/100/100/100,
        100/100/100/100/100/100,
        100/100/100/100/100/100,
        100/100/100/100/100/100,
        100/100/100/100/100/100,
        100/100/100/100/100/100,
        90/90/60/100/70/90
    }
\end{scope}

\begin{scope}[xshift=7.5cm, yshift=0cm]
    \drawheatmap{Basket horizontal offset $\bm{\rho}_b$ ($\lambda = 0.01m$)}
    {$-16\lambda$, $-8\lambda$, $-2\lambda$, $-1\lambda$, Base, $+1\lambda$, $+2\lambda$, $+8\lambda$, $+16\lambda$}
    {
        20/40/60/70/30/0,
        100/90/100/100/100/100,
        100/100/100/100/100/100,
        100/100/100/100/100/100,
        100/100/100/100/100/100,
        100/100/100/100/100/100,
        100/100/100/100/100/100,
        100/100/100/100/90/100,
        0/0/0/0/0/0
    }
\end{scope}

\begin{scope}[xshift=0cm, yshift=-7.0cm]
    \drawheatmap{Grasp z-offset $z_g$ ($\lambda = 0.005m$)}
    {$-16\lambda$, $-8\lambda$, $-2\lambda$, $-1\lambda$, Base, $+1\lambda$, $+2\lambda$, $+8\lambda$, $+16\lambda$}
    {
        100/100/100/70/100/100,
        100/100/100/100/100/100,
        100/100/100/100/100/100,
        100/100/100/100/100/100,
        100/100/100/100/100/100,
        100/90/100/100/100/100,
        100/90/100/100/100/100,
        100/100/100/80/100/100,
        60/70/40/30/80/100
    }
\end{scope}

\begin{scope}[xshift=7.5cm, yshift=-7.0cm]
    \drawheatmap{Place z-offset $z_p$ ($\lambda = 0.005m$)}
    {$-16\lambda$, $-8\lambda$, $-2\lambda$, $-1\lambda$, Base, $+1\lambda$, $+2\lambda$, $+8\lambda$, $+16\lambda$}
    {
        10/10/0/0/20/10,
        100/100/100/100/100/90,
        100/100/100/100/100/100,
        100/100/100/100/100/100,
        100/100/100/100/100/100,
        100/100/100/100/100/100,
        100/100/100/100/100/100,
        100/100/100/100/100/100,
        90/30/100/60/50/60
    }
\end{scope}

\end{tikzpicture} 
}
\vspace{-8mm}
\caption{Ablation studies on parameters across environments. Heatmap visualization indicates system boundary failures, with darker reds signaling higher failure rates.}
\label{fig:ablation-offset}
\vspace{-1em}
\end{figure}

In this ablation study, we test whether \ours{} depends on carefully tuned handoff parameters by varying the extraction radii ($r_o$, $r_b$) and the horizontal ($\bm{\rho}_o$, $\bm{\rho}_b$) and vertical target offsets ($z_g$, $z_p$) on the Alphabet Soup task. Figure~\ref{fig:ablation-radius} illustrates that success remains above 80\% across the radii we tested, so this limit lies outside our sweep; confirming it would require substantially larger extraction regions. Figure~\ref{fig:ablation-offset} additionally shows that sensitivity is parameter-dependent. For $\bm{\rho}_b$, $z_g$, and $z_p$, \ours{} remains at or near perfect success across $\pm8\lambda$; performance degrades only at $\pm16\lambda$, and collapses entirely for $\bm{\rho}_b$ at $+16\lambda$ and $z_p$ at $-16\lambda$. The horizontal grasp offset $\bm{\rho}_o$ is the tightest parameter: success is preserved within $\pm2\lambda$, degrades to 40–80\% at $-4\lambda$, and collapses to 0–50\% at $-8\lambda$. These results indicate a feasible handoff band that is broad but asymmetric, with $\bm{\rho}_o$ setting the binding constraint: the controller need only place the end effector within a geometrically feasible state covered by the local policy distribution, but that region is narrowest along the grasp-approach direction.

\section{Discussion and Future Work}
\label{sec:discussion}
Taken together, \textbf{RQ\#1--RQ\#4} show a consistent pattern: the hybrid factorization yields a modest reliability improvement on standard benchmarks but a large, broadly distributed robustness improvement once realistic perturbations are introduced, in both simulation and on a real robot. We discuss why this happens, why the current implementation is deliberately simple, and where it still fails.

\textbf{Why the factorization helps.}
Full-trajectory VLA training spreads supervision over many approach states that are easy to specify geometrically but expensive to cover with demos in unstructured scenes. The factorization helps for three reasons: (i) the transport controller maps diverse initial conditions and scene configurations into a compact local interaction distribution, reducing the distribution shift the policy must absorb. Moreover, transport is driven by an object-centric estimate rather than raw pixels; occlusion and appearance shift perturb only that estimate rather than the whole trajectory; (ii) shortening the VLA-controlled horizon limits the accumulation of perception and action errors; and (iii) demos concentrate on target objects and contact-sensitive actions rather than being diluted across repetitive approach motions, improving both data and inference efficiency. These mechanisms also delimit the benefit: as the extraction region expands, the local policy must recover an increasing share of long-range transport, and \ours{} progressively degenerates toward FullVLA---consistent with the upper bound of the feasible handoff band in Sec.~\ref{sec:ablation-study}. \ours{} and FullVLA are thus endpoints of a continuum, and the robustness gains arise from operating near the local end of it.

\textbf{Why the method is simple.}
The current implementation requires no additional component beyond the local policies: it requires no learned handoff-selection model and no modification of the action space, and target localization uses simulator objects and basket poses in LIBERO and off-the-shelf SAM3 segmentation on the real robot. Handoff targets are fixed offsets from these centers, and the transport controller and policies share the same 7D action format. This simplicity is a deliberate design choice, not an oversight: because the transport component is trivial, the robustness gain in Sec.~\ref{sec:robustness_results} is difficult to attribute to controller sophistication and is therefore most plausibly explained by the
factorization itself; fully isolating this attribution requires the decoupled ablations. It is also a strength for
benchmarking: if a proportional controller plus local VLA already closes most of the robustness gap, then full end-to-end training spends significant capacity on avoidable transport behavior.

\textbf{Limitations and Future Work.}
\ours{} depends on reliable target localization and reasonable handoff offsets: failures still occur if the center estimate is wrong, if the offset occludes the relevant affordance, or if the local policy is invoked outside its sphere-conditioned distribution. The implementation assumes access to object and basket centers, from the simulator state in LIBERO and calibrated object-centric targets in the real setup, and uses fixed offsets rather than selecting handoff position from perception, reachability, or policy confidence. LIBERO-Challenge injects visual and physical perturbations while preserving the official task predicate, but it does not yet model all real-contact effects of clutter. Future work should extend \ours{} to include learned online object localization, adaptive handoff selection, local recovery behaviors, collision-aware transport, and longer multi-object tasks in which the controller must sequence multiple local VLA skills.

\section{Conclusion}
We introduced \ours{}, a model-agnostic framework that factorizes object-centric manipulation into geometric transport and local VLA interaction. Across four VLA backbones, the proposed factorization preserves or improves performance on standard LIBERO tasks, while providing substantially larger gains under the clutter, distractors, obstruction, illumination variation, visual shift, and composed perturbations introduced by LIBERO-Challenge. It also improves demo efficiency, reduces VLA inference cost, and transfers effectively to a physical robot.

\textit{The central lesson is that end-to-end learning is not necessarily the most effective use of a VLA.} Free-space transport introduces many visually diverse but geometrically equivalent states, diluting supervision and extending the horizon over which errors accumulate. In contrast, semantic grounding and contact-rich local interaction are precisely the regimes where learned visuomotor policies provide the greatest value. \ours{} therefore assigns each component the subproblem it is best to solve: geometry handles structured and repeatable motion, while the VLA handles uncertain and contact-rich interaction. Our results suggest that robust and efficient robot foundation models may depend not only on scaling policies and datasets but also on the design of better boundaries between explicit structure and learned intelligence.

\bibliographystyle{class/IEEEtran}
\bibliography{class/IEEEabrv,class/reference}

\end{document}